\documentclass[pdflatex,sn-mathphys-num]{sn-jnl}% Math and Physical Sciences Numbered Reference Style
\usepackage{enumitem}

\newenvironment{promptblock}
{
\begin{quote}
\small\ttfamily\raggedright
}
{
\end{quote}
}
\usepackage{graphicx}%
\usepackage{multirow}%
\usepackage{amsmath,amssymb,amsfonts}%
\usepackage{amsthm}%
\usepackage{mathrsfs}%
\usepackage[title]{appendix}%
\usepackage{placeins}
\usepackage{xcolor}%
\usepackage{textcomp}%
\usepackage{manyfoot}%
\usepackage{booktabs}%
\usepackage{algorithm}%
\usepackage{algorithmicx}%
\usepackage{algpseudocode}%
\usepackage{listings}%
\usepackage[utf8]{inputenc}
\DeclareUnicodeCharacter{00B7}{\ifmmode\mathord{\cdot}\else\textperiodcentered\fi}
\theoremstyle{thmstyleone}%
\theoremstyle{thmstyletwo}%

\theoremstyle{thmstylethree}%

\begin{document}

\title[Article Title]{Can General-Purpose Generative AI Supply Training Data for Medical Vision Models? A Multimodal Retrospective Study}


\author[1,2]{\fnm{Jiachuan} \sur{Peng}}

\author[3]{\fnm{Kyle} \sur{Lam}}

\author[1]{\fnm{Jierui} \sur{Qu}}

\author[4]{\fnm{Lin} \sur{Li}}

\author[5]{\fnm{Haining} \sur{Zheng}}

\author[6]{\fnm{Bang} \sur{Zheng}}

\author[7,8]{\fnm{Aniruddha} \sur{Agarwal}}

\author[8,9]{\fnm{Zaki} \sur{Almallah}}

\author[2]{\fnm{Min} \sur{Chen}}

\author[2]{\fnm{Vicente} \sur{Grau}}

\author[1,10]{\fnm{Hagai} \sur{Rossman}}

\author[1,11]{\fnm{Eran} \sur{Segal}}

\author[1]{\fnm{Jianing} \sur{Qiu}} \email{Jianing.Qiu@mbzuai.ac.ae}

\affil[1]{\orgname{Biological and Life Sciences Division, MBZUAI}, \orgaddress{\country{UAE}}}

\affil[2]{\orgname{Department of Engineering Science, University of Oxford}, \orgaddress{\country{UK}}}

\affil[3]{\orgname{Department of Surgery and Cancer, Imperial College London}, 
\orgaddress{\country{UK}}}

\affil[4]{\orgname{OATML, Department of Computer Science, University of Oxford}, \orgaddress{\country{UK}}}

\affil[5]{\orgname{Department of Ultrasound, Peking University International Hospital}, \orgaddress{\country{China}}}

\affil[6]{\orgname{School of Public Health, Peking University}, \orgaddress{\country{China}}}

\affil[7]{\orgname{Eye Institute, Cleveland Clinic Abu Dhabi}, \orgaddress{\country{UAE}}}

\affil[8]{\orgname{Cleveland Clinic Lerner College of Medicine, Case Western Reserve University}, \orgaddress{\country{USA}}}

\affil[9]{\orgname{Integrated Surgical Institute, Cleveland Clinic Abu Dhabi}, \orgaddress{\country{UAE}}}

\affil[10]{\orgname{Pheno.AI}, \orgaddress{\country{Israel}}}

\affil[11]{\orgname{Department of Computer Science and Applied Mathematics, Weizmann Institute of Science}, \orgaddress{\country{Israel}}}

\maketitle
\section*{Abstract}
\textbf{Background} Access to well-annotated, diverse medical imaging datasets remains a major constraint in developing and evaluating vision-based medical artificial intelligence (AI). General-purpose generative image models may offer a scalable source of synthetic medical images, but their utility for training, subgroup augmentation, and model auditing remains uncertain. We evaluated Nano Banana, a general-purpose image generation and editing model, as a source of synthetic medical images for medical AI research.

\noindent\textbf{Methods}
In this retrospective multimodal study, we included datasets across four imaging modalities. For chest radiography, fundus photography, and CT, models trained exclusively on synthetic images were externally validated on real-world datasets and compared with models trained with real images. Carotid ultrasound models were developed using a private Human Phenotype Project (HPP) cohort, including subjects aged 40–75 years, and were tested with and without the augmentation of synthetic data. Clinicians rated generated and edited images for clinical plausibility on a 1–5 scale. Models were evaluated using AUROC and AUPR to assess discrimination performance, calibration plots were used to evaluate risk estimation, and decision curve analysis was used to assess clinical utility.

\noindent\textbf{Findings}
Synthetic-only models achieved AUROCs of 0·877 (95\% CI: 0·843–0·910) for pneumonia recognition (n=14863), 0·963 (0·948–0·975) for retinal detachment recognition (n=1149), and 0·711 (0·600–0·823) for renal calculus detection (n=4122), comparable to real-data-trained models. However, their calibration and the net benefit associated with model-assisted clinical decision-making were not consistently more favorable than those of real-data-trained models.
For synthetic-real mixed training, a carotid ultrasound model generated an ordinal vascular age index that stratified 633 external subjects, which reveals significant differences in BMI and glucose levels (both $p<0·01$) between subgroups, providing strengthened associations between vascular aging and cardiometabolic health. To stress-test model shortcut learning, adding a moderate tessellated pattern via the editing function to fundus photographs reduced the true-class probability for central serous chorioretinopathy from 0·996 to 0·822. Clinician-guided editing increased low-quality CT scan plausibility from 3·10/5 to 4·30/5.

\noindent\textbf{Interpretation}
Synthetic data can complement real medical data by supporting model training, improving feature learning and subgroup representation, and facilitating model testing through controlled image editing. However, its clinical value should be established based on more comprehensive evaluation with calibration, decision-curve analysis, and clinician review, beyond solely discrimination performance.

\section{Introduction}\label{sec1}

Large, annotated, and clinically diverse imaging datasets are foundational to medical AI, but they are difficult to obtain at scale. Medical images are often siloed across institutions, subject to privacy and governance constraints, and costly to label with clinical expertise~\cite{rieke2020future,webb2018deep}. Even when large datasets are available, they may underrepresent rare diseases, minority populations, atypical presentations, or specific acquisition protocols. These gaps can limit model development, external validation, and the evaluation of algorithmic performance across clinically important subgroups.

Synthetic medical images offer one possible response to these constraints. Generative models have been used to create clinically plausible radiographs~\cite{bluethgen2025vision}, ophthalmic images~\cite{qiu2024development}, and other biomedical data~\cite{valanarasu2025multimodal}, and synthetic data have been explored for augmentation, privacy-preserving data sharing, and robustness testing~\cite{wang2025self,bluethgen2025vision}. However, much of this work has relied on domain-specific generative models or specialized medical datasets. Less is known about whether widely accessible, general-purpose image-generation systems can produce medical images that are useful for training or evaluating medical AI models.

General-purpose generative image models introduce a different set of opportunities and risks. Because these models are trained on broad image-text corpora, they may encode medical visual concepts without being explicitly designed for clinical use~\cite{xie2025sana}. Their accessibility could lower barriers to early-stage medical AI development, particularly in settings where real annotated data are limited. At the same time, synthetic images may contain artifacts, encode stereotypes, fail to preserve clinical coherence, or resemble public benchmark data that may have been present during model training. Therefore, their value should be assessed not only by visual realism, but also by downstream model performance, subgroup utility, clinical plausibility, and failure modes under expert review.

In this work, we evaluated Nano Banana~\cite{comanici2025gemini}, a general-purpose image generation and editing model, as a tool for developing and auditing vision-based medical AI. We studied four imaging modalities: chest radiography, computed tomography (CT), fundus photography, and carotid ultrasound. We asked three questions. First, can synthetic data from a general-purpose generator be used to train AI models that generalize to real medical images? Second, can targeted synthetic augmentation improve performance in underrepresented real-world subgroups? Third, can generative image editing support controlled model auditing, stress testing, and clinician-guided refinement of synthetic images? By addressing these questions across multiple modalities and tasks, this work provides a practical case study of both the promise and the limitations of general-purpose synthetic medical imaging data.

\begin{figure}[!t]
\centerline{\includegraphics[width=\linewidth]{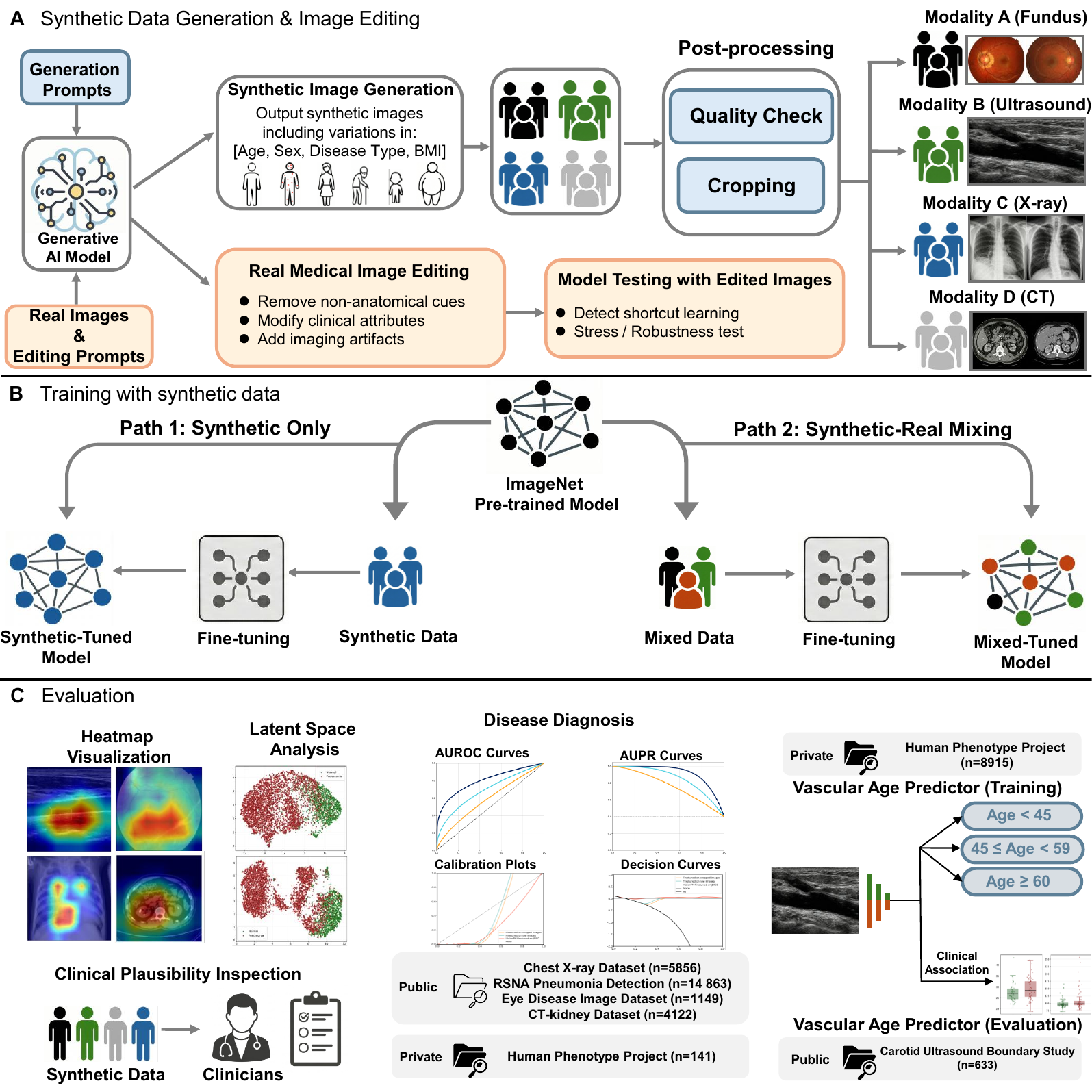}}
\caption{Overview of the proposed pipeline for developing and validating vision-based medical AI models with synthetic data generation and image editing using Nano Banana. (A) Synthetic Data Generation and Image Editing: prompts are used to generate synthetic medical images with variations in demographics, anatomical structures, modalities, and medical artifacts, and to edit real medical images by removing non-anatomical cues, modifying clinical attributes, or adding imaging artifacts for shortcut and robustness testing. Generated images undergo quality checks and cropping. (B) Model Training: models pre-trained on ImageNet are fine-tuned using two strategies: synthetic medical data only and a mix of synthetic and real-world data. (C) Model Evaluation: models are assessed quantitatively using discrimination metrics, calibration, clinical utility metrics, and qualitatively using heatmap visualization, latent space analysis, and clinical plausibility inspection. Experiments include both private and public datasets to ensure a fair evaluation of model generalization.}
\label{prompt}
\end{figure}

\section{Results}\label{sec2}

As shown in Fig.\ref{prompt}, we evaluated Nano Banana in three complementary roles: as a source of synthetic-only training data for diagnostic medical-imaging tasks, as a targeted augmentation tool for underrepresented subgroups in real-world data, and as an image-editing system for controlled model auditing and robustness testing. Synthetic images were generated across four imaging modalities: chest radiography, abdominal CT, fundus photography, and carotid ultrasound (Fig.\ref{prompt} A). The resulting images captured clinically relevant variation in disease status, anatomy, and demographic or anthropometric attributes, but raw outputs also contained recurrent artifacts, including digital watermarks, non-standard fields of view, and modality-specific border artifacts. Therefore, the primary analyses used post-processed synthetic images, with raw synthetic images retained for sensitivity analysis where appropriate. 

Across experiments, synthetic images were evaluated in two model-development settings (Fig.\ref{prompt} B). For pneumonia recognition, renal calculus detection, and retinal detachment (RD) detection, models were trained using synthetic images alone and externally tested on real-world datasets. For carotid ultrasound-based age prediction, targeted synthetic images were added to real Human Phenotype Project (HPP)~\cite{reicher2025deep} training data to test whether synthetic augmentation could improve performance in underrepresented subgroups (Fig.\ref{prompt} C). The trained model was then externally evaluated on the Carotid Ultrasound Boundary Study (CUBS) dataset~\cite{meiburger2021carotid} to assess the clinical associations between the carotid ultrasound-derived ordinal age index and cardiometabolic health. For clinical plausibility inspection, domain experts from each modality assessed the generated images using structured questionnaires and scored their realism and clinical plausibility. Across all tasks, expert readers rated each image using a 1–5 clinical plausibility scale. A score of 1 indicated that the image was not clinically plausible, 2 slightly plausible, 3 moderately plausible, and 4 highly plausible. A score of 5 indicated that the image was clinically plausible and could not be distinguished from a real medical image. In parallel, Nano Banana's image-editing capability was used to remove non-anatomical cues, introduce controlled pathological perturbations, and refine clinically implausible synthetic images. (Fig.\ref{prompt} A)

\begin{figure}[]
\includegraphics[height=0.8\textheight, keepaspectratio]{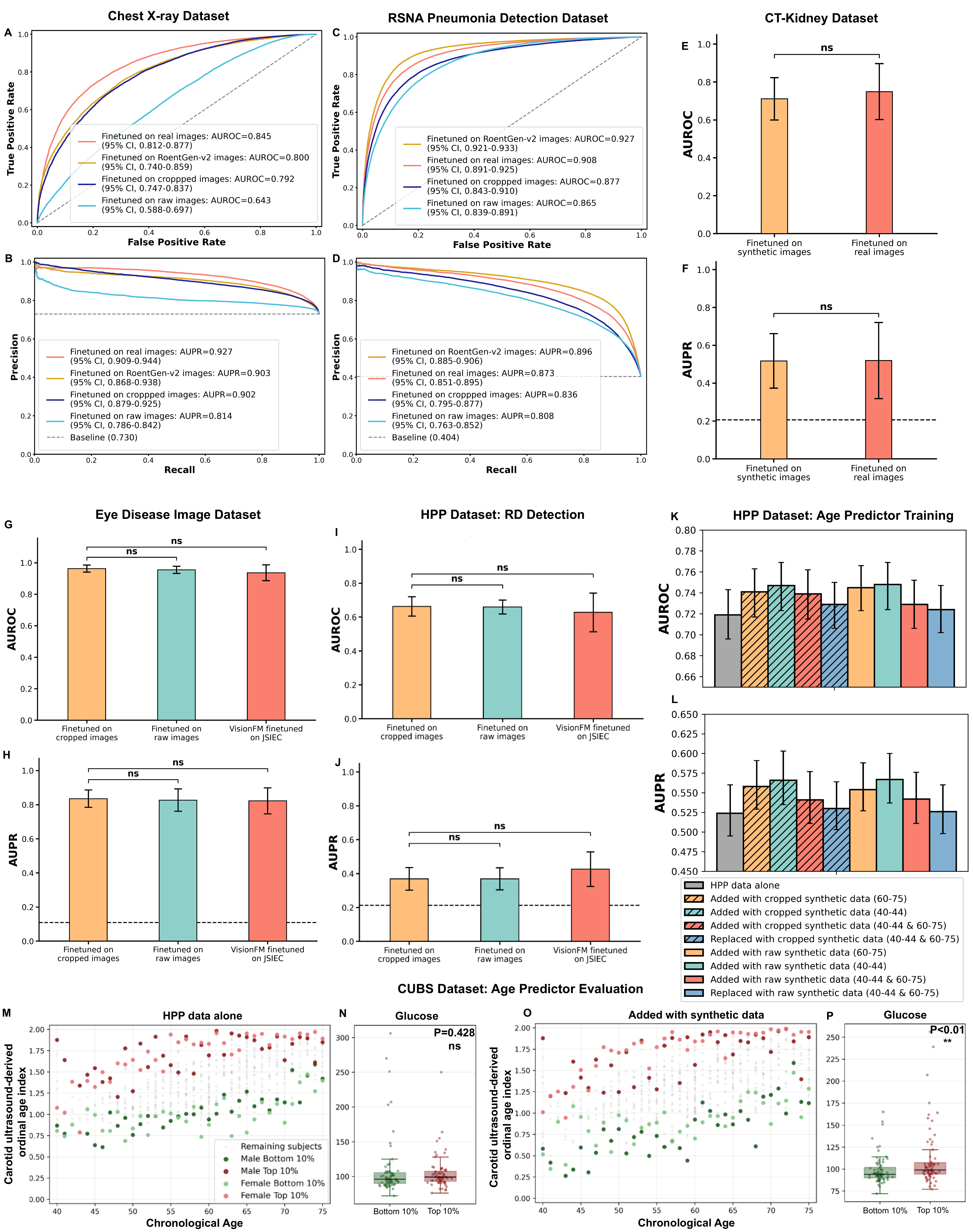}
\caption{Comprehensive evaluation of models fine-tuned with Nano Banana synthetic data across multiple medical imaging tasks. 
(A-D) AUROC and AUPR curves for disease recognition on the Chest X-ray Dataset (A-B) and RSNA Pneumonia Detection Dataset (C-D), comparing cropped and raw Nano Banana synthetic images with real-image and RoentGen-v2 baselines. 
(E-F) Performance comparison on the CT-Kidney Dataset between models fine-tuned using generated CT images and an equal number of real CT images for renal calculus detection.
(G-J) RD detection performance on the Eye Disease Image Dataset (G-H) and HPP Dataset (I-J), comparing cropped synthetic images, raw synthetic images, and VisionFM fine-tuned on JSIEC. 
(K-L) Results for HPP age prediction using HPP data alone or augmented with cropped/raw synthetic data across different age groups. 
(M–P) Age predictor external evaluation on the CUBS dataset. (M–N) Carotid ultrasound-derived ordinal age index grouping and glucose distributions using the HPP-only model. (O–P) Corresponding results using the synthetic data-augmented model. Bottom and top 10\% subgroups were selected.
Brackets denote paired statistical comparisons, and ns indicates not significant.}
\label{comb}
\end{figure}

\subsection{Synthetic-only training generalized to real-world diagnostic datasets}\label{syn_diagnosis}
Synthetic-only training produced models that generalized to real medical images across three diagnostic tasks: pneumonia recognition on chest x-ray (CXR) images, renal calculus detection on CT, and retinal detachment recognition on fundus photographs.

For pneumonia recognition, a classifier trained on cropped Nano Banana-generated chest radiographs achieved an AUROC of 0·792 (95\% CI: 0·747-0·837) and an AUPR of 0·902 (95\% CI: 0·879-0·925) on the Chest X-Ray dataset~\cite{kermany2018identifying}, while the same model trained on the same number of real chest radiographs from the CheXpert dataset~\cite{irvin2019chexpert} achieved an AUROC of 0·845 (95\% CI: 0·812-0·877) and an AUPR of 0·927 (95\% CI: 0·909-0·944) (Fig.~\ref{comb} A-B).
On the RSNA dataset~\cite{shih2019augmenting}, the cropped synthetic-image model achieved an AUROC of 0·877 (95\% CI: 0·843-0·910) and an AUPR of 0·836 (95\% CI: 0·795-0·877), compared with an AUROC of 0·908 (95\% CI: 0·891-0·925) and an AUPR of 0·873 (95\% CI: 0·851-0·895) for matched real-image training (Fig.~\ref{comb} C-D). The performance differences between models trained on cropped synthetic CXRs and those trained on matched real CXRs were not statistically significant on either external dataset. These findings suggest that the generated CXRs contained sufficient pneumonia-related imaging features to train a classifier that generalized to real clinical CXR datasets comparably to a classifier trained on real CXRs.

Post-processing markedly affected performance. Models trained on raw Nano Banana-generated chest radiographs, which included extended fields of view and digital watermarks, performed worse than those trained on cropped images. As shown in Fig.~\ref{comb} A-D, on the Chest X-Ray dataset, AUROC decreased from 0·792 (95\% CI: 0·747-0·837) to 0·643 (95\% CI: 0·588-0·697) with raw synthetic images, and AUPR decreased from 0·902 (95\% CI: 0·879-0·925) to 0·814 (95\% CI: 0·786-0·842). On the RSNA dataset, raw synthetic-image training achieved an AUROC of 0·865 (95\% CI: 0·839-0·891) and an AUPR of 0·808 (95\% CI: 0·763-0·852), lower than the corresponding cropped-image model. These findings show that clinical acquisition alignment and artifact removal were important determinants of synthetic-data utility.

Feature-space and saliency analyses supported the quantitative findings. Models fine-tuned on cropped Nano Banana chest radiographs produced more discriminative feature representations than models trained on raw synthetic images or ImageNet features alone (Supplemental Fig.~\ref{umap}). Grad-CAM visualizations showed attention over clinically plausible regions, including areas of consolidation in pneumonia cases and cardiothoracic structures in healthy controls (Supplemental Fig.~\ref{gradcam}).

To test whether these findings extended beyond chest radiographs, we next evaluated renal calculus detection on CT. A model trained on Nano Banana-generated CT images achieved an AUROC of 0·711 (95\% CI: 0·600-0·823) and an AUPR of 0·518 (95\% CI: 0·374-0·662) on the external CT-kidney dataset (Fig.~\ref{comb} E-F). A model trained on matched real CT images achieved an AUROC of 0·749 (95\% CI: 0·602-0·897) and an AUPR of 0·520 (95\% CI: 0·319-0·721). The performance differences were not statistically significant. As shown in Supplemental Fig. \ref{gradcam_ct}, Grad-CAM maps showed similar attention patterns for synthetic-trained and real-trained models, with activation over the renal region in stone-positive cases.

For RD detection, the model trained on cropped synthetic fundus photographs achieved an AUROC of 0·963 (95\% CI: 0·948-0·975) and an AUPR of 0·836 (95\% CI: 0·803-0·866) on the public Eye Disease Image Dataset. VisionFM~\cite{qiu2024development}, an ophthalmic foundation model, achieved an AUROC of 0·937 (95\% CI: 0·905-0·969) and an AUPR of 0·823 (95\% CI: 0·772-0·864), with no statistically significant difference between models (Fig.~\ref{comb} G-H). However, both models showed substantially lower performance on the private HPP retinal detachment cohort. The Nano Banana-trained model achieved an AUROC of 0·663 (95\% CI: 0·627-0·698) and an AUPR of 0·370 (95\% CI: 0·321-0·403), while VisionFM achieved an AUROC of 0·628 (95\% CI: 0·558-0·700) and an AUPR of 0·427 (95\% CI: 0·363-0·492) (Fig.~\ref{comb} I-J).

We additionally evaluated calibration and clinical utility using smoothed calibration plots and decision curve analysis~\cite{van2025evaluation}. Overall, models trained on synthetic and real images exhibit different patterns in these plots, despite their comparable discrimination performance. Concretely, the synthetic image-trained model showed a calibration curve (Supplemental Fig.~\ref{DCA} A, C) closer to the ideal diagonal line in the pneumonia and renal calculus recognition tasks, which suggests that its predicted probabilities were better aligned with the real-world risk distribution. It also indicates a reduced tendency of the model to underestimate or overestimate the true risk of both diseases. In the decision curves, the synthetic image-trained models in Supplemental Fig.~\ref{DCA} B remained above both reference strategies across the displayed threshold range. The treat-all (All) and treat-none (None) strategies assume intervention for all patients or no patients, respectively. Therefore, the higher model curves indicate additional positive net benefit from model-guided decisions. Moreover, in Supplemental Fig.~\ref{DCA} D, although neither model consistently exceeded both reference strategies, the synthetic image-trained model generally showed higher net benefit than the real image-trained model in the task of renal calculus recognition, suggesting improved relative clinical decision utility. However, results from other tasks and modalities showed that synthetic-only training was not consistently advantageous and generally tended to overestimate risk (Supplemental Fig.~\ref{DCA_supp}). These findings suggest that similar discrimination performance in terms of AUROC and AUPR does not ensure similar capability for risk estimation or clinical decision-making. It might depend on factors such as model architecture, training strategy, data scale, and task variation. Additional details, along with definitions of the calibration and clinical utility plots, are provided in Section~\ref{SI}.

\subsection{Targeted synthetic augmentation improved underrepresented subgroups}\label{US_improve}

We next evaluated whether targeted synthetic data could improve model performance in underrepresented subgroups of a real-world carotid ultrasound dataset. In the HPP age prediction task~\cite{reicher2024phenome}, the baseline model trained on HPP data alone achieved a mean AUROC of 0·719 (95\% CI: 0·696-0·743) and a mean AUPR of 0·524 (95\% CI: 0·495-0·560) (Fig.~\ref{comb} K–L). The original cohort was imbalanced, with fewer participants in the $\leq$ 44-year and $\geq$ 60-year groups than in the 45-59-year group. We therefore generated synthetic carotid ultrasound images targeting these underrepresented age ranges and added them only to the real-data training partition. 
As a result, we observed that synthetic augmentation improved age prediction performance across all tested augmentation settings (Fig.~\ref{comb} K–L). When synthetic data from both underrepresented age groups were added, subgroup performance improved for both the $\leq$ 44-year group and the $\geq$ 60-year group. For the $\leq$ 44-year group, AUROC increased from 0·748 (95\% CI: 0·715-0·781) to 0·777 (95\% CI: 0·745-0·805) and AUPR increased from 0·400 (95\% CI: 0·347-0·469) to 0·417 (95\% CI: 0·361-0·477). For the $\geq$ 60-year group, AUROC increased from 0·790 (95\% CI: 0·760 – 0·821) to 0·814 (95\% CI: 0·785 – 0·844) and AUPR increased from 0·467 (95\% CI: 0·407 – 0·532) to 0·504 (95\% CI: 0·445 – 0·569) (Supplemental Fig.~\ref{age_croppped} A–B). To control for the increase in training-set size introduced by augmentation, we further conducted a size-matched experiment, in which 240 real images were randomly removed from the $\leq$ 44-year and $\geq$ 60-year age groups (120 images each) and replaced with an equal number of synthetic images in the corresponding age group. Experimental results show that replacing real samples with cropped synthetic data achieved a mean AUROC of 0·729 (95\% CI: 0·706–0·750) and a mean AUPR of 0·530 (95\% CI: 0·503–0·564), while replacement with raw synthetic data achieved a mean AUROC of 0·724 (95\% CI: 0·702–0·747) and a mean AUPR of 0·526 (95\% CI: 0·498–0·560). Both were similar to the real data-only performance (AUROC: 0·719 and AUPR: 0·524). This suggested that at a fixed dataset size, replacing some real data with synthetic data does not harm performance, and when synthetic data are added on top of the real dataset, they boosted performance by augmenting effective sample size and diversity (the best performance was achieved when adding the 40–44 year group, increasing mean AUROC from 0·719 (95\% CI: 0·696 – 0·743) to 0·747 (95\% CI: 0·723 – 0·769) and mean AUPR from 0·524 (95\% CI: 0·495 – 0·560) to 0·566 (95\% CI: 0·535 – 0·603)). We also applied a similar subgroup augmentation strategy to BMI prediction and observed improved performance, which is detailed in Section~\ref{SI}.
In Supplemental Fig.~\ref{gradcam_US}, Grad-CAM analysis suggested that models augmented with cropped synthetic images attended to broader anatomically relevant regions, whereas models trained with raw images sometimes showed displaced attention. 

Furthermore, to investigate the potential clinical utility of the age prediction model, we applied both the HPP-only model and the model trained with synthetic-real mixed data to the external CUBS cohort. We selected the model augmented with synthetic samples from the 40–44 year age group because it achieved the best age prediction performance in the internal HPP evaluation. The models generated a carotid ultrasound-derived ordinal age index for each participant, which was used to characterize the individual’s model-inferred vascular age level. Within each individual chronological age, participants in the top (red) and bottom (green) 10\% of the score distribution were defined as the model-inferred vascularly older and younger subgroups, respectively (Fig.~\ref{comb} M and O). This stratification is also independent of the sex distribution. Darker and lighter shades indicate male and female participants, respectively. Further details of data processing, score calculation, and subgroup definition are provided in Section~\ref{SI}. Among all evaluated biomarkers and clinical measurements, BMI was significantly higher in the vascularly older subgroup for both models (p $<$ 0·01; Supplemental Fig.~\ref{measure_comp}). In addition, glucose was significantly higher in the vascularly older subgroup identified by the augmented model (p $<$ 0·01; Fig.~\ref{comb} P), whereas no significant difference was observed using the HPP-only model (p = 0·428; Fig.~\ref{comb} N). The significantly higher BMI and glucose levels observed in the vascularly older subgroup are consistent with previous evidence linking metabolic risk factors to vascular structure and function changes~\cite{seyedhoseinpour2023association,gomez2016association,scuteri2010central}. Notably, although both models identified a significant BMI difference, the glucose difference was detected only by the model with synthetic data augmentation. This finding suggests that synthetic augmentation enabled the model to learn more informative vascular features associated with cardiometabolic health, thereby improving its ability to distinguish individuals with different vascular health profiles independent of chronological age and sex. In the longer term, such model improvements enabled by targeted synthetic data augmentation may have the potential for the discovery of novel biomarkers and the identification of individuals at elevated cardiometabolic risk.

% The best overall performance was observed when adding synthetic images from the 40-44-year age group, increasing mean AUROC from 0·719 (95\% CI: 0·696-0·743) to 0·747 (95\% CI: 0·723-0·769) and mean AUPR from 0·524 (95\% CI: 0·495-0·560) to 0·566 (95\% CI: 0·535-0·603) (Fig.~\ref{comb} K–L). 

% Synthetic augmentation also improved BMI prediction. The baseline HPP-only model achieved an MAE of 1·852 (95\% CI: 1·774-1·936) and an $R^2$ of 0·339 (Fig.~\ref{comb} N). Prediction error was higher among participants with BMI $\leq$ 25, particularly those with BMI $<$ 22. After adding cropped synthetic carotid ultrasound images targeting BMI 17-25, overall MAE decreased to 1·799 (95\% CI: 1·726-1·880) and $R^2$ increased to 0·372 (Fig.~\ref{comb} M). Subgroup error improved for participants with BMI $\leq$ 25, with MAE decreasing from 1·910 to 1·845, while performance for participants with BMI $>$ 25 was not sacrificed; MAE also decreased in that group from 1·790 to 1·750.

% The benefit of synthetic augmentation depended on image processing. Raw synthetic BMI images did not improve overall performance, with MAE remaining 1·852 (95\% CI: 1·774-1·940) and $R^2$ decreasing to 0·331 (Supplemental Fig. \ref{age_croppped} C, middle). Although raw synthetic images reduced error in the BMI $\leq$ 25 subgroup, they worsened error in the BMI $>$ 25 subgroup. 

\subsection{Generative editing enabled model stress testing and clinician-guided refinement}\label{editing}

Beyond generating new images, Nano Banana was evaluated as an image-editing tool for model auditing. In chest radiographs, we first used editing to remove visible labels, side markers, and other external annotations from paired RSNA images while preserving underlying anatomy. A chest x-ray foundation model, Eva-x~\cite{yao2025eva}, achieved an AUROC of 0·965 (95\% CI: 0·960-0·970) and an AUPR of 0·974 (95\% CI: 0·970-0·978) on the original RSNA subset. After removal of non-anatomical cues, performance remained essentially unchanged, with AUROC of 0·967 (95\% CI: 0·963-0·972) and AUPR of 0·976 (95\% CI: 0·972-0·979). The small effect size suggests that this foundation model did not strongly rely on visible markers or labels as shortcuts in this evaluation set.

We then used image editing to introduce a subtle pathological perturbation by adding a small calcified granuloma to normal RSNA chest radiographs while leaving pneumonia images unchanged. This perturbation reduced AUROC from 0·965 (95\% CI: 0·960-0·970) to 0·957 (95\% CI: 0·951-0·963) and AUPR from 0·974 (95\% CI: 0·970-0·978) to 0·968 (95\% CI: 0·964-0·973) (p $<$ 0·001). Meanwhile, the mean predicted probability of the normal class among true-normal cases decreased from 0·754 (95\% CI: 0·727–0·781) to 0·721 (95\% CI: 0·693–0·748). In contrast, the mean normal class probability among true positive cases remained unchanged at 0·188 (95\% CI: 0·171–0·206), as expected because the pneumonia images were not modified. These degradations are most likely driven by the model’s predictions for edited normal cases shifting toward the pneumonia class, resulting in a less clear decision boundary and an increase in false positives. 

We then used image-editing to stress test eye disease recognition on the JSIEC dataset~\cite{cen2021automatic} containing 39 disease categories. We evaluated the ophthalmic foundation model, VisionFM~\cite{qiu2024development}, on the original test images and edited images. Adding shadow artifacts to fundus images produced little change in AUROC, which decreased from 0·994 to 0·992, but produced a larger decrease in AUPR, from 0·928 to 0·888. Mild tessellated-pattern editing produced limited changes in true-class probabilities, which reflect the model’s prediction confidence for the ground-truth disease category, across most categories. In contrast, moderate tessellation reduced model confidence for some diseases, including central serous
chorioretinopathy (0·996 to 0·822) and preretinal hemorrhage (0·983 to 0·857). These experiments show that generative editing can create controlled perturbation sets for evaluating model robustness under clinically plausible image-quality and background-pattern changes.

Finally, clinician-guided editing improved the plausibility of lower-quality synthetic CT images. Among 100 generated CT images reviewed by an expert, the overall plausibility score was 4·16/5, but stone-positive cases were less plausible than normal cases. Ten lower-quality images had a mean plausibility score of 3·10, with expert comments noting overly rounded or artificial-appearing stones. When these comments were used as targeted editing instructions, the mean plausibility score of the same images increased to 4·30. This suggests that image editing may support an iterative workflow in which clinicians identify failure modes and guide refinement of synthetic medical images.

\subsection{Expert review identified clinically plausible images and important failure modes}\label{expertread}

Expert review revealed an important distinction between image realism, attribute fidelity, and clinical coherence. Edited fundus images were generally judged clinically plausible. Across 95 edited fundus images, the mean plausibility score was 4·21/5, and 77 images received a score of 4 or 5. Tessellated-pattern edits were rated more plausible than shadow-artifact edits, with mean scores of 4·56 and 3·90, respectively.

Carotid ultrasound images were also generally plausible, but attribute fidelity varied by task. In the age prediction task, the expert correctly identified the older-appearing image in 66·7\% of evaluable pairs, with 14 of 50 pairs marked as ``cannot recognize". The corresponding mean clinical plausibility score was 3·82/5. In the BMI prediction task, the generated images received a mean plausibility score of 4·00/5. However, experts correctly identified the higher-BMI image in 18/50 pairs (36·0\%), incorrectly identified it in 16/50 pairs (32·0\%), and were unable to determine the BMI ordering in 16/50 pairs (32·0\%). Among the 34 evaluated pairs, the accuracy was 52·9\%. Thus,  despite their relatively high visual plausibility, the generated images did not consistently reflect the intended BMI differences. Although adding BMI-conditioned synthetic images was associated with modestly improved regression performance, it remains uncertain whether this improvement resulted from clinically meaningful BMI-related features or other image characteristics.

We also carried out an additional study to examine the clinical relevance of generated carotid ultrasound, i.e., carotid stenosis classification. The Doppler stenosis task exposed a more complex failure mode. The expert achieved 65·3\% four-class accuracy for stenosis grading, suggesting that the generated Doppler images contained partially interpretable stenosis-related information. However, this task received the lowest plausibility score, 3·47/5. The expert identified inconsistencies among B-mode anatomy, color Doppler flow, spectral waveform, velocity scale, sampling position, and expected hemodynamic pattern. An example of generated Doppler images with moderate stenosis is shown in Fig. \ref{doppler_US}.

\section{Discussion}\label{sec3}

In this multimodal case study, we evaluated whether images generated and edited by a general-purpose generative AI model could support vision-based medical AI development. Three findings emerged. First, models trained only on synthetic images generalized to real datasets for certain diagnostic tasks, including pneumonia recognition, renal calculus detection, and retinal detachment recognition, but showed varying calibration of predicted probabilities and clinical utility across tasks. Second, targeted synthetic augmentation improved age prediction in underrepresented carotid ultrasound subgroups and yielded a carotid ultrasound-derived ordinal age index with stronger associations with clinically relevant cardiometabolic measures in external validation. Third, generative image editing provided a practical way for auditing model behavior. Imaging artifacts could be selectively added or removed to assess model reliance on non-pathological cues, and controlled pathological perturbations can enable systematic stress testing and robustness evaluation. Moreover, clinician-guided editing allowed clinically implausible synthetic images to be iteratively refined.

These findings suggest that the value of synthetic medical images should not be judged by visual realism alone. Some synthetic images were distinguishable from real images (in a visual Turing test we conducted for the Nano Banana-generated chest radiographs, Supplemental Fig. \ref{turing}, the accuracy for a clinician to distinguish between real and fake images is 76\%), yet still contained disease-relevant features that supported downstream model training. Conversely, images that appeared plausible could fail to encode the intended attribute reliably, as seen in the BMI-conditioned carotid ultrasound task. Moreover, our expert review of generated Doppler ultrasound images suggests an important difference between component-level visual plausibility and case-level physiological coherence. Although the generator could reproduce recognisable appearances of B-mode, colour Doppler, and spectral Doppler ultrasound, these components were not always mutually consistent or aligned with the intended stenosis grade. For example, plaque morphology and luminal narrowing on B-mode imaging did not necessarily correspond to the expected colour-flow disturbance, spectral waveform, velocity measurements, or sampling configuration. Evaluation of synthetic medical image should therefore move beyond visual realism and prompt-attribute fidelity towards a multidimensional, physiology-constrained, case-level assessment that considers clinical plausibility, cross-component physiological coherence, acquisition alignment, and downstream model utility concurrently.

A key practical insight is that synthetic images required curation before they became useful. Raw generated images often contained digital watermarks, non-standard fields of view, or extraneous borders. These artifacts affected model performance, particularly in chest radiographs and carotid ultrasound. Post-processing improved alignment with real acquisition conventions and was associated with better generalization. This indicates that synthetic data pipelines should include image review, artifact removal, and modality-specific quality control rather than relying on prompt generation alone.

Image editing was a distinct capability of the general-purpose generative model. Beyond generating training data, the model could modify real images to test whether classifiers relied on non-anatomical markers, were sensitive to subtle pathology-related changes, or were robust to imaging artifacts. Removal of chest radiograph labels and markers had little effect on tested classifiers, whereas adding a small calcified granuloma shifted predictions toward pneumonia. 

The clinician-guided CT experiment further suggests a human-in-the-loop workflow for improving synthetic datasets. Expert comments identified implausible stone appearances, and targeted editing increased the clinical plausibility of selected images. This positions expert review not only as a screening step but also as an iterative refinement mechanism. In future work, clinicians could help define prompt constraints, identify recurrent artifacts, guide image editing, and establish exclusion criteria before synthetic images are used for model development.

Several limitations should temper the interpretation of these findings. The study included a limited number of modalities, tasks, and expert readers. The synthetic training sets were small, and the experiments were not designed to establish formal noninferiority between synthetic and real data. Another limitation is that the generated CXRs represented adults, whereas one external test set (i.e., Chest X-ray dataset) included a pediatric cohort. This creates data distribution mismatch between the synthetic training and evaluation data, which may limit the generalizability of the findings. There are also unresolved governance issues. Synthetic medical images may reduce dependence on identifiable patient data during early experimentation, but they do not eliminate the need for transparency, provenance tracking, bias assessment, and external validation. Moreover, general-purpose generators may encode stereotypical associations between clinical attributes and hence produce implausible anatomy. For example, when generating carotid ultrasound images for an older age group, an unconstrained prompt containing only age-related information may cause the model to introduce large plaques or severe stenosis, even when these pathological findings are not intended. Prompts should therefore be carefully designed to isolate the target attribute and explicitly constrain possible but unintended co-occurring features. Otherwise, synthetic datasets may contain generator-induced biases that are subsequently learned by downstream models, potentially reducing their generalizability to real clinical populations. Models developed using synthetic data must therefore be validated on representative real-world datasets and evaluated across clinically relevant subgroups before clinical deployment or regulatory consideration~\cite{lam2026physicians}.

\section{Conclusion}\label{sec4}
% In this study, we investigated the utility of synthetic medical imaging data generated by Nano Banana for developing AI models. While the external validation results are promising, further exploration is required across wider medical domains, modalities and tasks to demonstrate the value of synthetic medical data generated by industrial models. Beyond the technical challenges, complementary policies and guidelines for transparent and responsible use of synthetic medical data delivered through Internet-based interfaces should be developed in tandem.

General-purpose generative models could expand the range of toolkit available for medical AI development. They can rapidly create synthetic training data, augment the representation of specific sub-populations, and support controlled image-editing experiments to test what visual cues an AI model relies on. However, our findings also show that synthetic medical images cannot be assumed to be accurate or clinically meaningful simply because they appear realistic. Their utility depends on careful curation, review by clinical experts, consistency with real imaging protocols, and validation using independent real-world patient data. Synthetic data should therefore be viewed as a complement to, rather than a replacement for, real-world medical imaging data.

% \section*{Data sharing}
% The publicly available datasets used in this study can be accessed from their respective original repositories or data providers, as cited in the manuscript, subject to their corresponding terms of use.
% The Human Phenotype Project (HPP) data used in this study are not publicly available owing to privacy and ethical considerations set forth by the study's institutional review board. HPP data are accessible to bona fide researchers at universities and other research institutions through the HPP data access procedure (https://humanphenotypeproject.org/data-access). Interested researchers should contact info@pheno.ai. Following approval and completion of a data use agreement permitting non-commercial research use, access is provided through a secure cloud-based environment. The raw data used in this study were verified by more than one academic-only author (J.C.P., J.R.Q., and J.N.Q.).

\FloatBarrier
\newpage
\section{Supplementary materials}\label{SI}
\setcounter{figure}{0}
\renewcommand{\thefigure}{S\arabic{figure}}
\renewcommand{\theHfigure}{S\arabic{figure}}

\subsection{Methods and Supplementary Results}
As illustrated in Fig.\ref{prompt} A, we first prompted Nano Banana to generate synthetic data across four modalities. Across all modalities, the generated images contained Nano Banana’s digital watermark. Most synthetic CXRs exhibit a field-of-view extending beyond the thoracic region, while some carotid ultrasound images show extraneous white borders. We therefore applied modality-specific post-processing to align the synthetic data with standard acquisition protocols and remove artifactual regions. Specifically, we cropped the lower 30\% of each CXR, removed 10\% from both the top and bottom borders of each fundus image, and cropped 10\% from the right border and 3\% from the top, bottom, and left borders of each carotid ultrasound image. For CT images, we used simple digital image processing method to remove the watermarks since they locate at black background areas. Supplemental Fig. \ref{examples} shows comparisons between the original images generated by Nano Banana and their processed counterparts. In addition to synthetic data generation, we used Nano Banana’s image editing function to modify real medical images, including removing non-anatomical cues, modifying clinical attributes, and adding imaging artifacts. These edited images were used as a probing tool to evaluate potential shortcut learning in AI models and to assess their robustness. 

Next, we train the models either solely on synthetic data for pneumonia, renal calculus, and RD recognition or on mixed synthetic and real-world data for age prediction and ordinal age index derivation (Fig.\ref{prompt} B), and evaluated on real-world cohorts with both quantitative and qualitative analysis (Fig.\ref{prompt} C). 
All images used in this study were resized to 224$\times$224 pixels and normalized using the standard ImageNet~\cite{deng2009imagenet} mean and standard deviation. Performance was evaluated using Area Under the Receiver Operating Characteristic curve (AUROC) and the Area Under the Precision-Recall curve (AUPR). BMI regression was evaluated via the Coefficient of Determination ($R^2$) and the Mean Absolute Error (MAE). 95\% Confidence Interval (CI) was calculated via non-parametric bootstrapping with 1000 iterations. The overall statistical analysis was performed using a paired t-test. Welch t-tests and Fisher’s exact tests were used in Section~\ref{US_improve} specifically.

\subsection*{Pneumonia Recognition from Chest X-rays}

For CXRs, we synthesized 300 frontal-view CXRs that varied in demographics, anthropometrics, imaging positions (posteroanterior and anteroposterior views), and pathological features, and constructed a balanced dataset with a 50:50 ratio of pneumonia and healthy subjects, which was then partitioned into training (n=220), validation (n=30), and test (n=50) sets.
To assess model performance, we conducted external validation using two real-world public datasets for pneumonia recognition: the Chest X-Ray dataset (5856 anteroposterior CXRs; 1583 healthy, 4273 pneumonia) \cite{kermany2018identifying} and the 2018 RSNA Pneumonia Detection Challenge dataset (14 863 images; 8851 normal, 6012 lung opacity) \cite{shih2019augmenting}. For the RSNA dataset, images labeled as ‘Lung Opacity’ were categorized as pneumonia cases and ‘Normal’ images as healthy controls. Images labeled ‘No Lung Opacity / Not Normal’ were excluded. Meanwhile, not all opacity cases represent pneumonia, and they might reflect other abnormalities, such as interstitial changes or nodules. Therefore, we acknowledge that this labeling strategy may introduce diagnostic errors. Also, to ensure robust and holistic external evaluation, we merged the predefined training and test partitions into a single evaluation set for these real-world datasets.

We leveraged a ResNet-50, pre-trained on the ImageNet dataset, as the base classifier for CXRs. We then attached a two-layer multi-layer perceptron (MLP) and fine-tuned the network on our curated synthetic dataset for 20 epochs with a fixed learning rate of $5 \times 10^{-5}$. Augmentations were applied to the training images, including random resized cropping, affine transformation, horizontal flipping, and rotation. The best-performing model on the internal CXR data, attaining 0·994 AUROC (95\% CI: 0·973 - 1·000) and 0·995 AUPR (95\% CI: 0·984 - 1·000) on the test set, was selected for external validation on the two real-world datasets. We further incorporated a real-data training control to enable a direct comparison under the same experimental setting. Specifically, we randomly sampled a balanced subset of real CXRs from CheXpert \cite{irvin2019chexpert} with the same sample size as the synthetic training set, and trained the model using the same architecture, hyperparameters, and training protocol. Both the synthetic-data and real-data training experiments were repeated ten times with different random seeds.

In addition to the results presented in Section \ref{syn_diagnosis}, we also curated 300 synthetic CXRs, including 150 healthy cases and 150 pneumonia cases, from a publicly available dataset generated by RoentGen-v2~\cite{moroianu2025improving}, a state-of-the-art text-to-image diffusion model for CXR generation. We then fine-tuned another ResNet-50 on the RoentGen-v2 synthetic CXRs using the same fine-tuning protocol. As shown by the yellow curves in Fig.~\ref{comb} (A-D), RoentGen-v2-based training achieved performance comparable to our cropped synthetic-image training on the Chest X-Ray dataset, while showing better performance on the RSNA dataset.

\begin{figure}[H]
\centerline{\includegraphics[width=\linewidth]{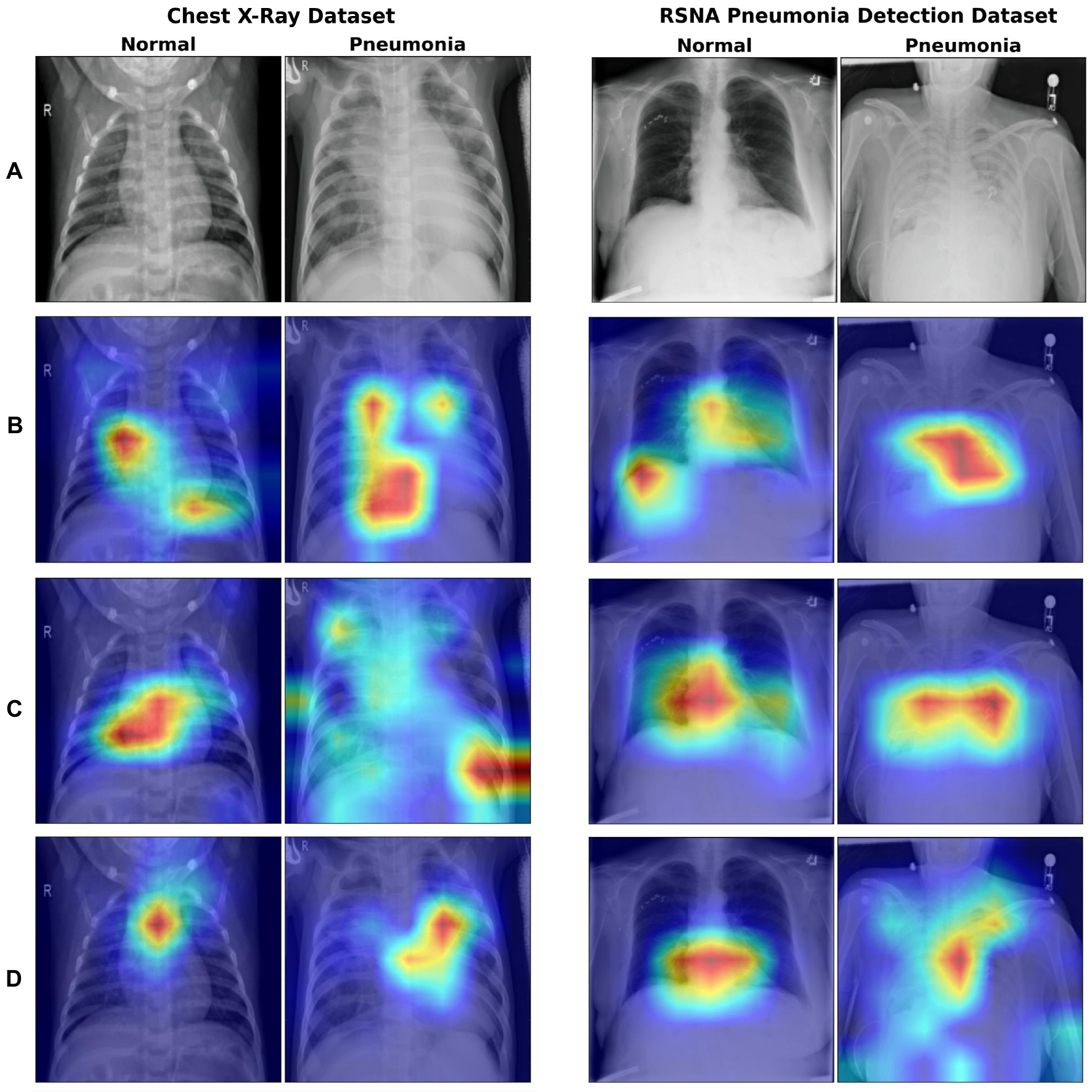}}
\caption{Grad-CAM visualization of the feature map of the last convolutional layer before the classification head. A: original image; B: fine-tuned on cropped CXRs generated by Nano Banana; C: on raw Nano Banana-generated CXRs; and D: on CXRs generated by RoentGen-v2.}
\label{gradcam}
\end{figure}

Moreover, as briefly summarized in the main text, we further evaluated whether the performance gain from synthetic CXRs was associated with more discriminative feature representations rather than features already learned from ImageNet. Here, we provide the detailed feature extraction, clustering, quantitative evaluation, and UMAP visualization results. First, we pre-extracted features of images in the two public datasets using the ResNet-50 fine-tuned with Nano Banana's cropped synthetic CXRs, with raw synthetic CXRs generated by Nano Banana, with RoentGen-v2's synthetic CXRs, and the ResNet-50 pre-trained on ImageNet without fine-tuning. We then applied K-means to perform unsupervised feature clustering and obtained the class predictions. As K-means produces hard labels independent of threshold values, we assess the unsupervised classification using Accuracy (Acc). Furthermore, we include the adjusted Rand index (ARI) to evaluate how well the clustering results match the true data distribution. Additionally, we employed UMAP \cite{mcinnes2018umap} to reduce the dimensionality of the extracted features and visualize them in two-dimensional spaces.
As shown in Supplemental Fig.~\ref{umap}, the ResNet-50 fine-tuned on cropped synthetic CXRs generated by Nano Banana (A, E) produced more discriminative feature representations, showing less overlapping and intertwining between clusters. Furthermore, it achieves higher Acc and ARI scores on both datasets, corroborating the efficacy of the Nano Banana generated synthetic CXRs. By contrast, although the model fine-tuned on raw images (B, F) and the non-fine-tuned model (D, H) exhibited different patterns in the UMAP visualizations, the former showed no significant advantage in the quantitative metrics predicted by K-means.
We also used Grad-CAM~\cite{selvaraju2017grad} to visualize feature maps, highlighting the areas where classifiers trained with different synthetic data focus on. For the classifier fine-tuned on cropped Nano Banana CXRs (Supplemental Fig.~\ref{gradcam} B), Grad-CAM indicated dominant attention over clinically salient features including areas of consolidation within pneumonia cases and cardiac and diaphragmatic borders within healthy cases.

In addition to image generation, we investigated whether Nano Banana could be used as an image editing tool to assess model sensitivity to non-anatomical visual cues. Here, we provide details of the procedures corresponding to the CXR-related results reported in Section~\ref{editing}.
We first constructed an RSNA evaluation subset containing 100 pneumonia and 100 normal CXRs. To obtain the classifiers used for this editing analysis, we adapted the Eva-x foundation model \cite{yao2025eva} using balanced CheXpert training subsets. The backbone was downloaded from the publicly available repository and kept frozen during training, while only the classification head was fine-tuned. Specifically, across 10 random seeds, we randomly sampled 1000 pneumonia and 1000 normal images from CheXpert for training each classifier. Nano Banana editing was then applied only to the RSNA evaluation subset to remove visible labels, side markers, and other external image annotations while preserving the underlying anatomy. Representative examples before and after editing are shown in Supplementary Fig. \ref{cxr_edit} (row 1).
We then evaluated the trained Eva-x based classifiers on the paired original and edited RSNA images. Results are summarized in Section~\ref{editing}

% On the original RSNA subset, the classifiers achieved an AUROC of 0·965 (95\% CI: 0·960--0·970) and an AUPR of 0·974 (95\% CI: 0·970--0·978). After removal of these non-anatomical cues, performance remained similar, with a slight increase to an AUROC of 0·967 (95\% CI: 0·963--0·972) and an AUPR of 0·976 (95\% CI: 0·972--0·979). Although the performance change was statistically significant, the magnitude of the change was very subtle, suggesting that the classifier did not strongly rely on non-anatomical markers or labels as shortcuts. Meanwhile, it indicates that Nano Banana maintained high image fidelity after image editing.

We next examined whether a subtle pathology-related cue, rather than a purely non-anatomical imaging marker, could influence model predictions. To this end, we used Nano Banana to add a small calcified granuloma only to the normal images in the same RSNA evaluation subset, while leaving all pneumonia images unchanged. We also show examples in Supplementary Fig. \ref{cxr_edit} (row 2) Using the same frozen Eva-x model and training protocol, we found performance drop on the edited set as detailed in Section~\ref{editing}. 

% The AUROC decreased from 0·965 (95\% CI: 0·960–0·970) on the original subset to 0·957 (95\% CI: 0·951–0·963) after editing, and the AUPR decreased from 0·974 (95\% CI: 0·970–0·978) to 0·968 (95\% CI: 0·964–0·973) (p $<$ 0·001). Meanwhile, the mean predicted probability of the normal class among true-normal cases decreased from 0·754 (95\% CI: 0·727–0·781) to 0·721 (95\% CI: 0·693–0·748). In contrast, the mean normal class probability among true positive cases remained unchanged at 0·188 (95\% CI: 0·171–0·206), as expected because the pneumonia images were not modified. These degradations are most likely driven by the model’s predictions for edited normal cases shifting toward the pneumonia class, resulting in a less clear decision boundary and an increase in false positives.

\subsection*{Renal Calculus Detection from Computed Tomography Imaging}
This section describes the procedures used for the renal calculus detection experiment with CT imaging presented in Section~\ref{syn_diagnosis}. Specifically, we generated a same class-balanced synthetic kidney CT dataset consisting of 150 images with renal calculi and 150 images without renal calculi, with visible watermarks removed (training: n=220; validation: n=30; test: n=50). To provide a matched comparison with real data, we also sampled real kidney CT images from the Axial CT Imaging Dataset \cite{abdalla2025kidney}, using 150 images per class in each run. All experiments were repeated with five random seeds. The trained models were externally evaluated on an independent CT-Kidney Dataset (4122 images; 3274 normal, 848 renal calculus) \cite{islam2022vision}, for which only axial CT slices were retained to better align the imaging plane distribution between the training and test sets.
We used a DenseNet-161 backbone and followed the same data augmentation and training protocol as in the CXR experiments. 

% As shown in Fig.~\ref{comb} (E-F), the model trained on Nano Banana-generated CT images achieved a mean AUROC of 0·711 (95\% CI: 0·600--0·823) and a mean AUPR of 0·518 (95\% CI: 0·374--0·662) on the external CT-Kidney test set. In comparison, the model trained on real CT images achieved a mean AUROC of 0·749 (95\% CI: 0·602--0·897) and a mean AUPR of 0·520 (95\% CI: 0·319--0·721). The performance differences between them were not statistically significant, suggesting that Nano Banana-generated CT images can provide comparable utility to real CT images in this task.

We further used Grad-CAM to compare the feature activation patterns of models trained on Nano Banana-generated CT images and matched real CT images. As shown in Supplemental Fig.\ref{gradcam_ct}, the two models exhibited highly similar attention patterns. In renal calculus cases, both models primarily highlighted the renal region containing the stone, while in normal cases the attention was more diffuse over the abdominal CT field.

Finally, we provide details of CT image editing results presented in Section \ref{editing}. We randomly sampled 50 images from each class, including normal and renal calculus cases, and invited a domain expert to evaluate their clinical plausibility. The 100 generated images achieved an overall clinical plausibility score of 4·16/5. However, stone cases were less plausible than normal cases (3·40 vs. 4·92). The expert provided comments for 10 lower quality or problematic images, whose mean score was only 3·10, with comments mainly noting overly rounded or fake-looking stones. We then used these expert comments as targeted instructions for the model’s image editing function to refine the same 10 images, followed by a second round of clinical plausibility assessment. After editing, we observe a significant improvement, with the mean score increasing to 4·30.

% Overall, these results demonstrate that the image editing function can serve as an iterative, clinician feedback-guided refinement tool, which could help correct clinically implausible visual artifacts and improve the realism of synthetic CT images.

\subsection*{Retinal Detachment Detection from Fundus Photographs}

% \begin{figure}[!t]
% \centerline{\includegraphics[width=\linewidth]{figure/performance_comp_rd.pdf}}
% \caption{Performance comparison between the models fine-tuned on raw fundus images generated by Nano Banana (light blue), on cropped Nano Banana images (dark blue). AUROC (A, C) and AUPR (B, D) are reported. We tested the model on the public Eye Disease Image Dataset (n=1149) and the private HPP Dataset (n=141).}
% \label{supervised_rd}
% \end{figure}

We also provide details for RD recognition based on color fundus photographs. We generated 45° center-view fundus images comprising 150 healthy eyes and 150 eyes with retinal detachment. The side of the eye (left or right) was randomly determined by Nano Banana. We composed a class-balanced RD dataset randomly split into training (n = 220), validation (n = 30), and test (n = 50) sets, following the same scheme as the synthetic CXR dataset. We included the public Eye Disease Image Dataset (1149 images; 1024 normal, 125 RD) \cite{rashid2024eye} and the restricted-access, private HPP dataset (141 images; 111 normal, 30 RD) \cite{reicher2025deep} for external validation.

In this experiment, we used a DenseNet-161 backbone pre-trained on ImageNet. We adopted a richer set of augmentations, including random resized cropping, affine transformation, rotation, color jitter, Gaussian noise, and Gaussian blurring. 
% To mitigate overfitting, we fine-tuned for
% % only 2 epochs on the severe RD subset 
% only 1 epoch on the composed RD dataset. 
After fine-tuning, the best-performing model achieved 
0·995 AUROC (95\% CI: 0·981 – 1·000) and 0·995 AUPR (95\% CI: 0·752 – 1·000) on the internal test set. As a comparison with real data-trained model, we further adapted a foundation model VisionFM \cite{qiu2024development} by fine-tuning it on the Normal and Rhegmatogenous RD categories from the JSIEC dataset \cite{cen2021automatic}.
% whereas on the severe RD dataset, the model reached 0·995 AUROC (95\% CI: 0·981 – 1·000) and 0·995 AUPR (95\% CI: 0·752 – 1·000).

% mild: an AUROC of 0·901 (95\% CI: 0·803 – 0·974) and an AUPR of 0·881 (95\% CI: 0·752 – 0·978)

We then directly applied the two models to recognize RD on the public Eye Disease Image Dataset and on the private HPP dataset. In addition to the results in Section \ref{syn_diagnosis}, we also observed no statistically significant difference between models trained on cropped and raw synthetic images in this task. This might imply that watermarks located outside the fundus imaging field may have limited influence on model generalization. 
Meanwhile, the comparable results achieved by the synthetic data-trained model and the foundation model on both public and private datasets indicate the prospect of synthetic data for eye disease diagnosis. These results also suggest that the lower performance on the private dataset is not specific to models trained with generated images, but more likely reflects the intrinsic difficulty of disease recognition on this private dataset. Nonetheless, the performance drop from public to private data should be interpreted carefully, as public medical data might have been included in the training data of Nano Banana.

% \textcolor{red}{Despite the heterogeneity in imaging quality and views, the model trained on cropped synthetic images achieved a mean AUROC of 0·963 (95\% CI: 0·948--0·975) and a mean AUPR of 0·836 (95\% CI: 0·803--0·866) on this real-world dataset (Fig. 2 G-H). Meanwhile, VisionFM achieved a mean AUROC of 0·937 (95\% CI: 0·905--0·969) and a mean AUPR of 0·823 (95\% CI: 0·772--0·864), showing no statistical significance for the performance difference between the two models.}

% We further tested them on the HPP cohort. \textcolor{red}{VisionFM achieved a mean AUROC of 0·628 (95\% CI: 0·558–0·700) and a mean AUPR of 0·427 (95\% CI: 0·363–0·492), while the Nano Banana-trained model finetuned on cropped images achieved a mean AUROC of 0·663 (95\% CI: 0·627–0·698) and a mean AUPR of 0·370 (95\% CI: 0·321–0·403) (Fig.~\ref{comb} I-J). The differences between these three were not statistically significant. 

% \textcolor{blue}{Although the generated and public fundus images are not visually the same, this may indicate a cascade effect, i.e., AI models developed with synthetic data generated by models developed with Internet data may tend to perform well on the public datasets. (DO we still need this sentence??)}

\begin{figure}[H]
\centerline{\includegraphics[width=\linewidth]{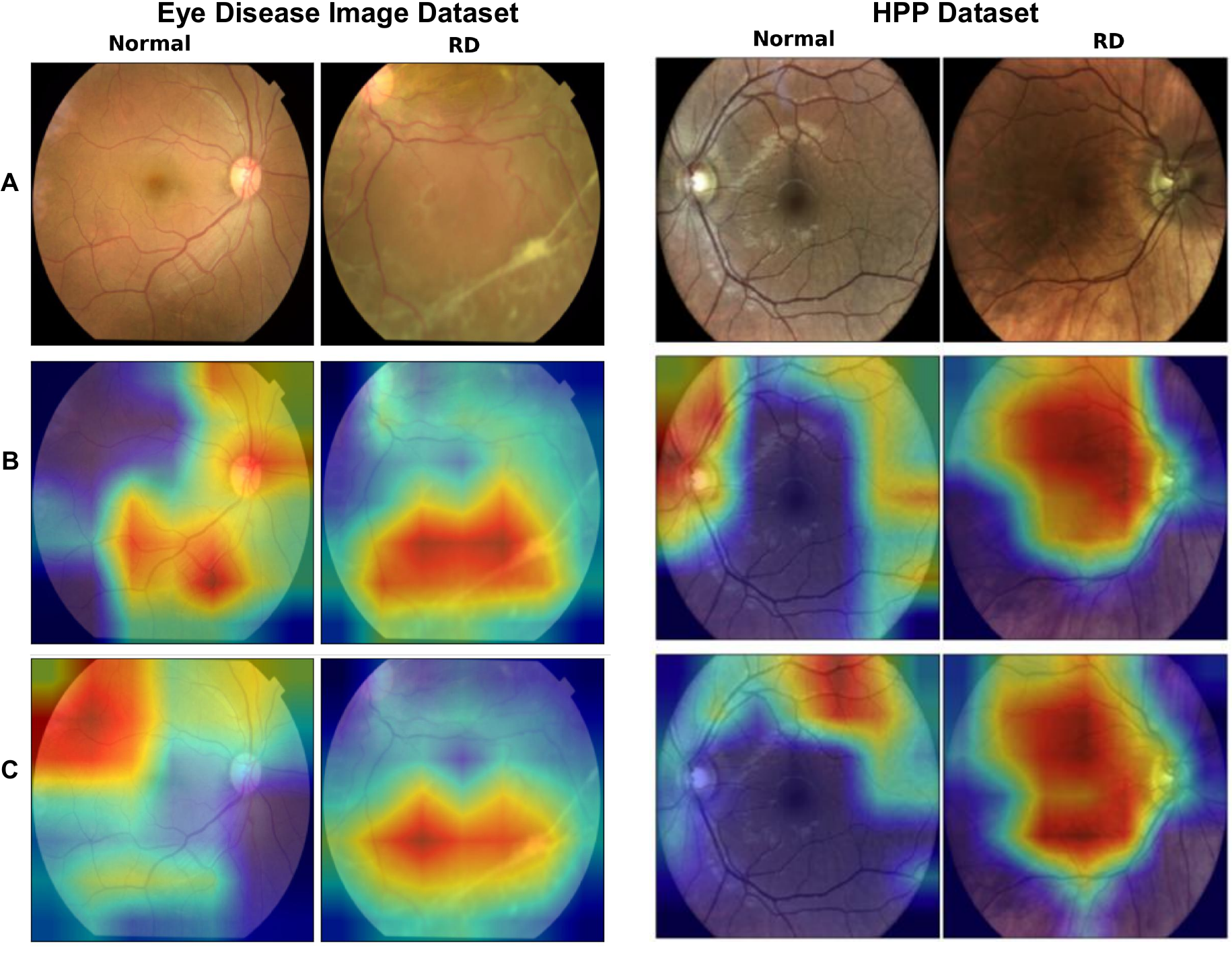}}
\caption{Grad-CAM visualization of the feature map of the last convolutional layer before the classification head. A: original image; B: fine-tuned on cropped Nano Banana-generated fundus images; and C: on raw Nano Banana-generated fundus images.}
\label{gradcam_rd}
\end{figure}

Supplemental Fig.~\ref{gradcam_rd} visualizes the feature maps of the model. Cases with clear RD exhibited concentrated attentions over the detached retinal regions in the fundus image (column 2), whereas for subtle RD or normal eyes, the feature maps primarily highlighted the retinal vasculature and optic disc region.

\begin{figure}[H]
\centerline{\includegraphics[width=\linewidth]{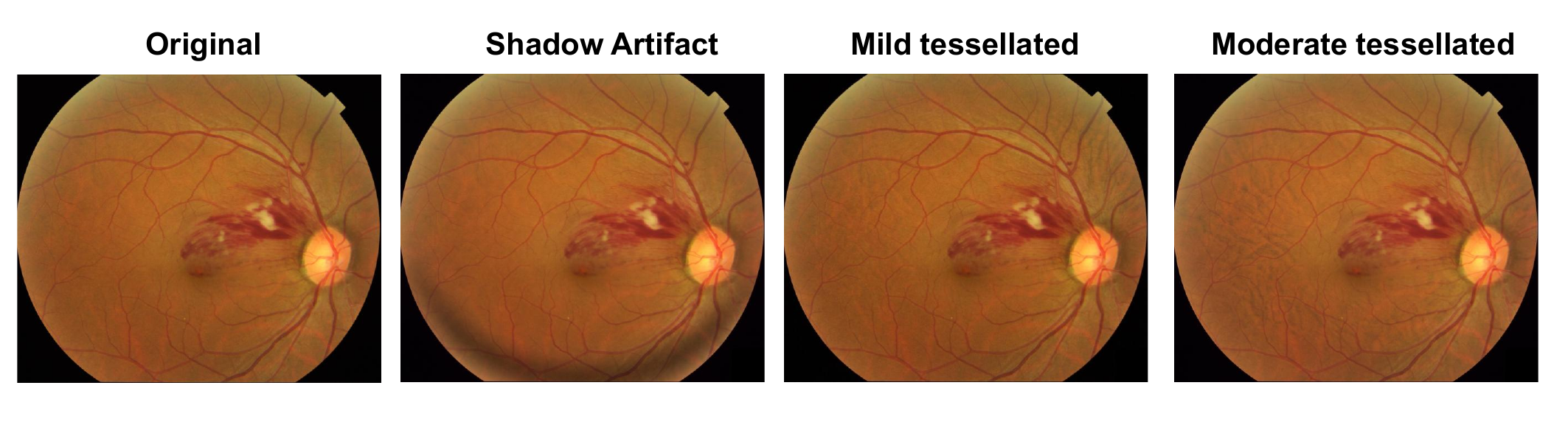}}
\caption{Example images demonstrating Nano Banana editing function. From left to right: the original image sampled from JSIEC , edited shadow artifact image, edited mild tessellated fundus image, and edited moderate tessellated fundus image.}
\label{fundus_edit}
\end{figure}

Finally, we provide additional details on the fundus image-editing results presented in Section~\ref{editing} and the expert reader study described in Section~\ref{expertread}. For fundus image editing, using the official VisionFM checkpoint fine-tuned on the JSIEC dataset, which contains 39 disease categories, we evaluated the model on the original test images and on edited images with added imaging shadow artifacts (Supplemental Fig.~\ref{fundus_edit}, column 1-2). 
After shadow editing, the AUROC remained nearly unchanged, decreasing slightly from 0·994 to 0·992, whereas the AUPR showed a larger decrease from 0·928 to 0·888.
Furthermore, we applied tessellated-pattern editing to seven disease categories with relatively clear primary lesions, including branch retinal vein occlusion, central retinal vein occlusion, retinal artery occlusion, central serous chorioretinopathy, macular hole, massive hard exudates, and preretinal hemorrhage. Representative editing results are shown in Supplemental Fig.~\ref{fundus_edit} (column 3-4). Under mild tessellation, the model remained generally stable, with true-class probabilities showing little changes for branch retinal vein occlusion (0·930 to 0·935), central retinal vein occlusion (1·000 to 1·000), retinal artery occlusion (1·000 to 0·999), central serous chorioretinopathy (0·996 to 0·996), massive hard exudates (0·870 to 0·859), and preretinal hemorrhage (0·983 to 0·981), while the probability for macular hole increased from 0·427 to 0·530. Under moderate tessellation, more obvious decreases were observed for central serous chorioretinopathy (0·996 to 0·822) and preretinal hemorrhage (0·983 to 0·857), with a smaller decrease for retinal artery occlusion (1·000 to 0·980). By contrast, branch retinal vein occlusion and central retinal vein occlusion remained stable, and massive hard exudates decreased only slightly from 0·870 to 0·857. These results suggest that the model was relatively robust to mild tessellated perturbations, whereas moderate tessellation imposed a stronger stress test for selected disease categories. Together, these experiments demonstrate the potential of Nano Banana’s image editing function as a tool for stress testing medical imaging models under targeted perturbations.

For the expert reader study, the expert evaluated 95 images in total, including 50 edited with imaging shadow and 45 with tessellated patterns. The overall clinical plausibility score was 4·21/5. Overall, 77/95 images (81·1\%) received a score of 4 or 5. For the imaging shadow task, the mean score was 3·90, with 36/50 images (72·0\%) scoring 4 or higher. For the tessellated pattern editing task, the mean score was higher at 4·56, with 41/45 images (91·1\%) scoring 4 or higher. Overall, these results suggest that the edited images were generally considered clinically plausible by the expert reader, with particularly high plausibility for tessellated pattern editing.

\subsection*{Calibration Plots and Decision Curve Analysis}
In this section, we provide additional information on calibration plots and decision curves, including their definitions and supplementary results. In a calibration plot, better performance is indicated by closer alignment with the diagonal line. In decision curve analysis, a model provides clinical benefit when its net benefit curve lies above the treat-none and treat-all strategies. The decision curves  show the complete 0-1 threshold range for exploratory comparison, while clinically meaningful threshold ranges should ultimately be prespecified by clinicians for each modality and diagnostic task.

\begin{figure}[H]
\centerline{\includegraphics[width=\linewidth]{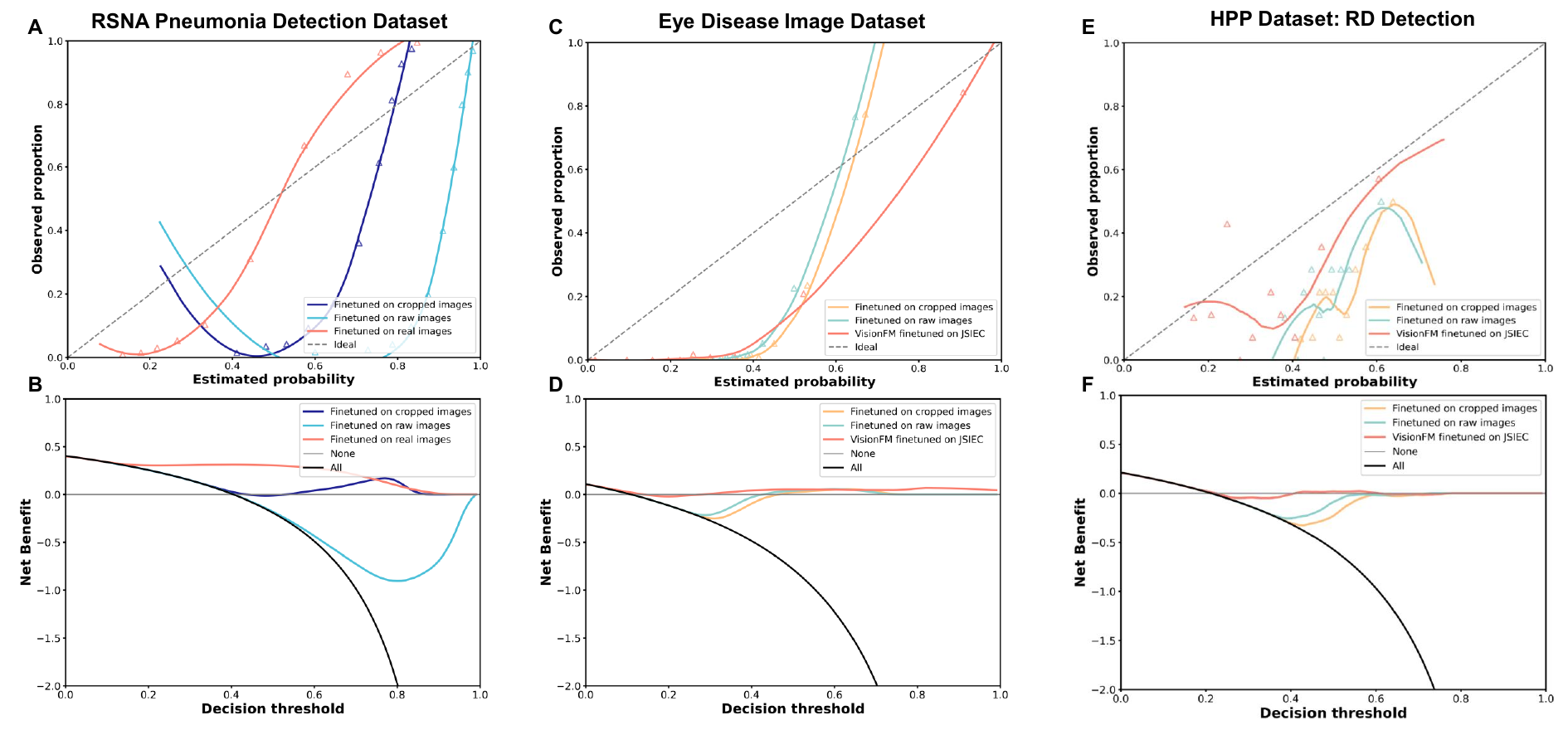}}
\caption{Additional calibration and decision-curve analyses on external chest-radiograph and fundus datasets.
Smoothed calibration plots (A, C, and E) and decision curves (B, D, and F) are shown for the RSNA Pneumonia Detection dataset (A-B), the Eye Disease Image Dataset (C-D), and the HPP retinal detachment cohort (E-F).}
    \label{DCA_supp}
\end{figure}

Additional external evaluations further showed that these relationships were task dependent (Supplemental Fig.~\ref{DCA_supp}). On the RSNA dataset (Supplemental Fig.~\ref{DCA_supp} A-B), the real image-trained model generally showed better calibration and net benefit than the synthetic image-trained models, particularly compared with training on raw generated images. In the two fundus cohorts (Supplemental Fig.~\ref{DCA_supp} C-F), VisionFM also showed improved calibration or decision curve performance across several threshold ranges. Notably, raw net benefit is prevalence dependent and should therefore be interpreted within each cohort rather than compared across datasets. For a perfect model, the maximum attainable raw net benefit equals the outcome prevalence. Consequently, the relatively low net-benefit scale in the Eye Disease Image Dataset partly reflects the low prevalence of RD rather than necessarily indicating poor model performance (Supplemental Fig.~\ref{DCA_supp} D). Also, the performance difference between VisionFM and synthetic-only models is expected, as VisionFM is a large ophthalmic foundation model with domain-specific pretraining, whereas the synthetic fundus classifiers used a substantially simpler DenseNet architecture trained on only 300 generated images. In addition, the repeated experiments were not fully symmetric in training-set diversity. More specifically, the same fixed synthetic image set was reused across seeds, whereas a different size-matched subset of real images was randomly sampled for each seed, giving the real-image condition access to a broader pool of image variation across repeated experiments. The aim of these analyses was therefore not to show that synthetic-only training universally outperformed real-image training, but to characterize the capabilities and limitations of general-purpose generated medical images. Overall, synthetic-only models achieved discrimination close to real image-trained or foundation-model baselines in several settings, but calibration and decision curve analyses showed additional task-specific differences that were not apparent from AUROC and AUPR alone.

\subsection*{Age and BMI Prediction from Carotid Ultrasound Images}

% The CXRs and fundus experiments revealed that models trained with synthetic images generated by Nano Banana could generalize to real-world data. 

We also investigated whether synthetic data could improve fairness and accuracy of a medical AI model by augmenting underrepresented real-world data. Therefore, this section provides additional details of results in Section \ref{US_improve}.
We used carotid ultrasound images from the HPP dataset for BMI (6820 images; 5456 training, 341 validation, 1023 test) and age prediction (8915 images; 7132 training, 445 validation, 1338 test). The BMI prediction cohort comprised participants with BMI $\leq 30$, including a normal-weight group $\mathrm{BMI}\leq 25$ (n=3556) and an overweight group with $\mathrm{BMI} > 25$ (n=3264) as defined by WHO~\cite{consultation2000obesity}. Among individuals with $\mathrm{BMI}\leq25$, the distribution is skewed toward the upper normal range, with 2378 (66·9\%) subjects having BMI between 22 and 25, and only 1178 (33·1\%) subjects having BMI $<$ 22 (17·3\% of all data). The age prediction cohort is categorized into three groups: $\leq44$ year (n = 1674), 45–59 year (n = 5484), and $\geq60$ year (n = 1757) \cite{skeet1983protecting}. We did not further split the $\leq44$ age group because the vast majority of subjects in this group are between 40 and 44 years old. All age and BMI values were measured at the time of image acquisition. We split the data into training, validation, and test sets using stratified sampling, ensuring that each split contained approximately the same proportions of subjects across age and BMI groups. We used DenseNet-161 as the backbone with the CXR augmentation scheme, but removed affine transformations and added color jitter, and trained the model for 10 epochs with the same learning rate. 

For HPP internal validation, using HPP data alone, the baseline age prediction model achieved a mean AUROC of 0·719 (95\% CI: 0·696 – 0·743) and a mean AUPR of 0·524 (95\% CI: 0·495 – 0·560) (Fig.~\ref{comb} K–L).
% with relatively lower AUPRs in the $\leq44$ year (0·400, 95\% CI: 0·347 – 0·469) and $\geq60$ year (0·467, 95\% CI: 0·407 – 0·532) groups than in the majority 45–59 year group (0·706, 95\% CI: 0·671 – 0·742) (Supplemental Fig.~\ref{age_croppped} A–B).
As data imbalance was observed in the original HPP cohort, which motivated the investigation of whether synthetic data generated for the targeted underrepresented populations could improve the accuracy, in our case, the $\leq44$ year (18·8\% of all data) and $\geq60$ year (19·7\% of all data) subgroups. Especially, as noted above, most subjects in the $\leq44$ year group of the HPP dataset are 40–44 years old. Accordingly, we synthesized 240 images (120 subjects aged 40–44 years and 120 subjects aged 60–75 years). All generated carotid ultrasound images were mixed into the training partition of the HPP real data for the training purpose only.

Specifically, we augmented the training set under three conditions: 1) with 120 generated images for the 40-44 year old group; 2) with 120 generated images for the 60–75 year old group, or 3) with both sets, and examined how these interventions affected performance on the test set. As summarized in Fig.~\ref{comb} K–L, incorporating synthetic images improved both metrics across all three augmentation settings. The best performance was achieved when adding the 40–44 year group, increasing mean AUROC from 0·719 (95\% CI: 0·696 – 0·743) to 0·747 (95\% CI: 0·723 – 0·769) and mean AUPR from 0·524 (95\% CI: 0·495 – 0·560) to 0·566 (95\% CI: 0·535 – 0·603). 

\begin{figure}[H]
\centerline{\includegraphics[width=\linewidth]{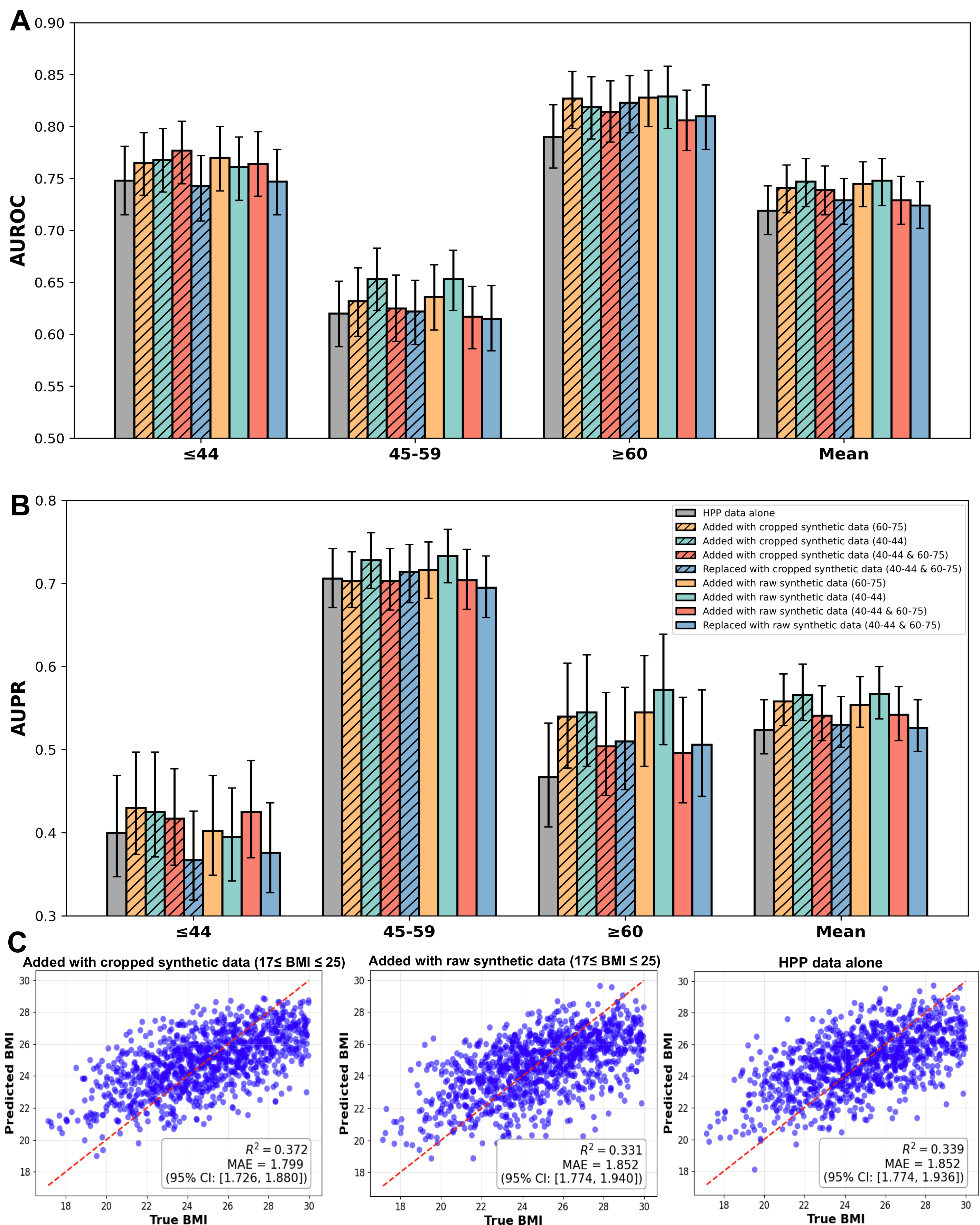}}
\caption{Results of three-class age prediction ($\leq44$, 45–59, and $\geq60$).
(A) AUROC and (B) AUPR evaluated across the three age groups and the overall mean. Each bar represents a different training configuration, including the baseline model trained with HPP data alone and models augmented with either cropped or raw synthetic images generated for specific age groups (40-44, 60-75, or both). Additional replacement experiments were performed by randomly deleting real images from each of the $\leq44$ and $\geq60$ groups and adding an equal number of cropped or raw synthetic images, thereby keeping the total training set size unchanged. Error bars indicate 95\% CI. (C) shows the BMI prediction results for HPP data added with cropped synthetic data (left), added with raw synthetic data (middle), and HPP data alone (right).}
\label{age_croppped}
\end{figure}

More importantly, we also observed performance improvement for each individual subgroup (Supplemental Fig.~\ref{age_croppped} A–B). When synthetic data from both underrepresented subgroups were added, performance improved for both subgroups, with AUROC/AUPR increasing from 0·748 (95\% CI: 0·715 – 0·781)/ 0·400 (95\% CI: 0·347 – 0·469) to 0·777 (95\% CI: 0·745 – 0·805) / 0·417 (95\% CI: 0·361 – 0·477) for the $\leq44$ year group and increasing from 0·790 (95\% CI: 0·760 – 0·821)/ 0·467 (95\% CI: 0·407 – 0·532) to 0·814 (95\% CI: 0·785 – 0·844) / 0·504 (95\% CI: 0·445 – 0·569) for the $\geq60$ year group. For the fixed dataset size experiment, replacement with cropped synthetic data achieved an AUROC/AUPR of 0·743 (95\% CI: 0·709–0·772)/0·367 (95\% CI: 0·319–0·426) for the $\leq44$-year group and 0·823 (95\% CI: 0·794–0·849)/0·510 (95\% CI: 0·452–0·575) for the $\geq60$-year group. Similarly, replacement with raw synthetic data achieved an AUROC/AUPR of 0·747 (95\% CI: 0·715–0·778)/0·376 (95\% CI: 0·328–0·436) and 0·810 (95\% CI: 0·778–0·840)/0·506 (95\% CI: 0·444–0·572), respectively. Compared with the HPP-only baseline, AUROC in the $\leq44$-year group was largely preserved, whereas performance in the $\geq60$-year group improved consistently for both cropped and raw synthetic replacement. Together with the comparable or slightly improved overall mean performance under a fixed training-set size, these results indicate that the gains observed in the augmentation experiments cannot be explained solely by the increased number of training samples. Instead, they support that the targeted synthetic images provide useful age-related information.

For external validation of our age predictor, we used the Cyprus cohort from the CUBS dataset because it contained the most complete set of relevant measurements. Participants outside the 40-75 year age range were excluded, resulting in 633 subjects. Each subject had paired left and right carotid ultrasound images. We independently input the left and right views into the age prediction models, and averaged the predicted class probabilities across the two views. Because the model predicted three age intervals, we defined a continuous carotid ultrasound-derived ordinal age index by assigning ordinal weights of 0, 1, and 2 to the $\leq44$-, 45-59, and $\geq60$ year classes, respectively:
\[
\mathrm{Ordinal\ Age\ Index}
=0 \times p_{\leq 44}+1 \times p_{45\text{--}59}+
2 \times p_{\geq 60}.\]
To identify participants with relatively high or low carotid ultrasound-derived ordinal age index, subjects were stratified jointly by chronological age and sex. Within each chronological age stratum, the subjects with the highest and lowest 10\% scores were assigned to the vascularly older and younger groups, respectively. We rounded up the selected number of subject to ensure that each group contained at least one subject in sparse strata. This was applied to male and female subjects individually to ensure matched chronological age and sex distributions between the two selected subgroups. This stratification procedure minimized confounding by chronological age and sex, yielding 71 participants in each subgroup.

In addition to the main results (Section~\ref{US_improve}), we further showed the BMI distributions for both models in Supplemental Fig.~\ref{measure_comp}. BMI was significantly higher in the vascularly older group for both the HPP-only and synthetic-data-augmented models (both p $<$ 0·01). Glucose was also significantly higher in the vascularly older group identified by the augmented model (p $<$ 0·01), whereas no significant difference was observed using the HPP-only model (p = 0·428). These results suggest that adding synthetic data improved age prediction performance and strengthened the model’s ability to capture clinically meaningful cardiovascular characteristics. The resulting carotid ultrasound-derived ordinal age index therefore provided clearer separation between the model-inferred vascularly younger and older groups. The more significant difference for glucose levels identified by the augmented model further indicates its potential to facilitate novel biomarker discovery.

We also performed synthetic augmentation for BMI prediction. The baseline model trained on HPP data alone yielded a MAE of 1·852 (95\% CI: 1·774 – 1·936) and an $R^2$ of 0·339 (Supplemental Fig.~\ref{age_croppped} C), with higher error rates for subjects with $\mathrm{BMI}\leq25$ (MAE = 1·910) than for those with $\mathrm{BMI}>25$ (MAE = 1·790). Furthermore, prediction errors were substantially higher in the BMI $<$ 22 subgroup (MAE = 2·710) compared with the 22–25 subgroup (MAE = 1·517).  Therefore, we investigated whether the model would improve the prediction accuracy for the subgroup of lower accuracy with synthetic data intervention. Specifically, we synthesized 100 images for subjects with BMI between 17 and 25, with the majority (64\%) falling in the 17-22 range, to intentionally augment the underrepresented BMI $<$ 22 subgroup in HPP data. We then mixed them with the training split of real data. 

Following this synthetic data intervention, we observed that the overall performance of BMI prediction was improved, with the MAE reducing to 1·799 (95\% CI: 1·726 - 1·880) and the $R^2$ increasing to 0·372 (Supplemental Fig.~\ref{age_croppped} C left). Further stratifying the result reveals that the model augmented with cropped synthetic data reduced the MAE for $\mathrm{BMI}\leq25$ (1·845 versus 1·910) while not sacrificing the MAE of  $\mathrm{BMI}>25$, which was also reduced (1·750 versus 1·790). Compared to HPP data alone (Supplemental Fig.~\ref{age_croppped} C right), adding cropped synthetic data pulls many peripheral points closer to the ideal fit line, indicating that the augmentation is effective in correcting large prediction errors.

Notably, after synthetic data intervention, the MAE decreases from 2·710 to 2·664 for subjects with $\mathrm{BMI}<22$ and decreases from 1·517 to 1·441 for subjects with $\mathrm{BMI}$ between 22 and 25 on the test set. 
% This indicates the effectiveness of improving the prediction for the targeted subgroup ($\mathrm{BMI}<22$), while not introducing extra bias for other subgroups ($22\leq\mathrm{BMI}\leq25$ and $\mathrm{BMI}>25$). 
By contrast, using raw synthetic images shows no overall improvement, with a MAE of 1·852 (95\% CI: 1·774 – 1·940) and a $R^2$ of 0·331 (Supplemental Fig. \ref{age_croppped} C, middle). Nonetheless, it reduced the MAE of $\mathrm{BMI}\leq25$ subgroup to 1·754 (2·369 for BMI $<$ 22 and 1·450 for $22\leq\mathrm{BMI}\leq25$), but resulted in a higher MAE (1·957) for the $\mathrm{BMI}>25$ subgroup. These results highlight the potential of synthetic data to mitigate biases in real-world datasets and the importance of appropriate post-processing of generated synthetic images.

As shown in Supplemental Fig.~\ref{gradcam_US}, Grad-CAM for age prediction shows predominantly local attention along the arterial far wall, consistent with age-related artery changes. In contrast, BMI prediction relies on more global patterns, with feature maps spanning the subcutaneous tissue. Notably, models trained with unprocessed raw synthetic images may exhibit displaced attentions (e.g., second column of Supplemental Fig.~\ref{gradcam_US} C), whereas augmentation with cropped images could extend attentions to broader anatomical regions. 

Complementary to the results presented in the main text (Section~\ref{expertread}), we provide additional details on the expert reader study on carotid ultrasound images. we invited an ultrasound imaging expert to perform a blinded reader evaluation. Three reader tasks were designed to assess the clinical plausibility and attribute fidelity of the generated images. In the first task, the reader was shown paired B-mode carotid ultrasound images generated with older and younger age prompts and was asked to identify which image appeared older, with an additional `cannot recognize' option. In the second task, the reader was shown paired B-mode carotid ultrasound images generated with different BMI prompts and was asked to identify which image corresponded to the higher BMI, again with a `cannot recognize' option. In the third task, the reader evaluated carotid ultrasound Doppler images generated to represent four stenosis categories, including no stenosis, mild stenosis, moderate stenosis, and severe stenosis.

The reader study showed that the generated carotid ultrasound images were generally considered clinically plausible, with mean clinical plausibility scores of 3·82/5 for the age task and 4·00/5 for the BMI task. For the age comparison task, after excluding `cannot recognize' responses, the reader achieved an accuracy of 66·7\%, with 14/50 pairs marked as `cannot recognize'. This suggests that the age-conditioned images contained some recognizable age-related anatomical variations. In contrast, the BMI comparison task showed weaker attribute recognizability, with an accuracy of 52·9\% after excluding `cannot recognize' responses, 16/50 pairs marked as `cannot recognize'. For the Doppler stenosis grading task, the reader achieved a four-class accuracy of 65·3\%, indicating that the generated Doppler images contained partially interpretable stenosis-related information. However, this task received the lowest clinical plausibility score, with a mean score of 3·47/5. The reader comments suggested that the main limitations were inconsistencies among the B-mode anatomy, color Doppler flow, spectral Doppler waveform, velocity scale, sampling position, and expected hemodynamic pattern.

\newpage

\subsection{Prompts}

We used the following prompts to generate synthetic medical images using Nano Banana.

\begin{enumerate}[label=\textbf{\arabic*)}, leftmargin=*]

\item Synthetic chest X-ray images:
\begin{promptblock}
Generate 10 Chest X-ray imaging data consisting of images representing different
[healthy/pneumonia] patients, generated individually as separate downloadable files
so I can download them one by one. The views you generate need to maintain a
consistent format, meaning the overall image is portrait-oriented, and the images
must show variations in gender, height, weight (fat and thin), and age, as well as
human posture and body stance during imaging, along with differences in lung texture
to ensure clinical authenticity and individual diversity.
\end{promptblock}

\item Synthetic fundus images:
\begin{promptblock}
Generate 10 fundus images representing different mild retinal detachment patients.
The views you generate need to maintain a consistent format, and the images must
show variations in gender, height, weight (fat and thin), and age, along with
differences in pathological texture to ensure clinical authenticity and individual
diversity. Each image must be presented on a separate square canvas.
\end{promptblock}

\item Synthetic carotid ultrasound images:
\begin{promptblock}
Generate 10 carotid ultrasound images representing various clinical cases comprising
mostly healthy subjects alongside few cases with conditions. The views must maintain
a consistent format and exhibit variations in gender, height, and weight. The age
range is restricted to [40--44/60--75] years old to capture the physiological watershed
around age [44/60], or the BMI range is restricted to 17--25. Do avoid stereotypical
depictions of severe plaque or occlusion in all cases.
\end{promptblock}

\item Synthetic CT images:
\begin{promptblock}
Generate exactly 5 separate synthetic axial non-contrast abdominal CT images as independent downloadable square image files, without creating a collage, grid, montage, contact sheet, or combined panel. Each image should have a 1:1 aspect ratio, with the axial CT slice centered on a square canvas, preserving the black background around the body contour like exported clinical CT images. Do not stretch the body to fill the canvas, and do not crop the body at any edge. The images should resemble realistic clinical non-contrast renal CT / CT KUB scans for [control cases / renal calculus cases], using grayscale soft-tissue abdominal windowing with a moderately dark, clinically realistic tone. There should be no IV contrast, CTA appearance, nephrographic enhancement, excretory contrast, opacified renal pelvis, or bright contrast-filled vessels. Kidneys should appear as medium-gray soft tissue, with dark gray fat, black air, medium-gray soft tissues, and very bright cortical bone only. Add realistic CT noise, mild grain, slight reconstruction blur, and natural partial-volume effects, making the images look like real DICOM CT slices exported as clinical grayscale images rather than polished illustrations. Across the 5 images, vary the anatomical slice level along the cranio-caudal axis, including upper renal pole, mid-kidney, renal hilum or renal pelvis, lower renal pole, and ureteropelvic junction or proximal ureter levels. Maintain anatomical consistency for each level, with upper slices showing more liver, spleen, stomach, or upper abdominal organs, mid slices showing fuller renal anatomy, renal hilum slices showing renal sinus fat, vessels, and collecting-system region, and lower slices showing smaller or partial kidneys with more bowel loops or proximal ureteral region. Each image should show realistic axial abdominal anatomy at or near the kidney, renal pelvis, or proximal ureter level, including anatomically plausible kidneys when visible, renal sinus fat, collecting-system region, vertebral body, psoas muscles, abdominal wall, bowel loops, retroperitoneal fat, and partial liver or spleen depending on slice level. Not every slice needs to show both kidneys equally; vary right-kidney dominance, left-kidney dominance, and bilateral kidney visibility naturally. Renal calculus status should follow this option: [non-stone: do not generate any renal calculi, urinary-tract stones, punctate hyperdense foci, stone-like calcifications, artificial dots, round markers, bright vessel-like structures, hydronephrosis, obstructing stones, or secondary signs of urinary obstruction / stone: generate clinically plausible renal calculi as small hyperdense foci within the renal calyx, renal pelvis, ureteropelvic junction, or proximal ureter, while avoiding artificial markers, random non-urinary calcifications, contrast-filled vessels, or exaggerated obstruction unless explicitly requested]. Vary patient body habitus, renal size, renal shape, cortical thickness, renal sinus fat prominence, renal pelvis visibility, collecting-system conspicuity, bowel gas, bowel contents, retroperitoneal fat, psoas muscle appearance, CT noise, sharpness, reconstruction texture, and mild grain while keeping the anatomy clinically plausible
\end{promptblock}

\end{enumerate}

We show representative raw and cropped CXRs, fundus images, carotid ultrasound images, and abdominal CT images, generated by Nano Banana in Supplemental Fig.~\ref{examples}.

\begin{figure}[H]
\centering{\includegraphics[height=0.82\textheight, keepaspectratio]{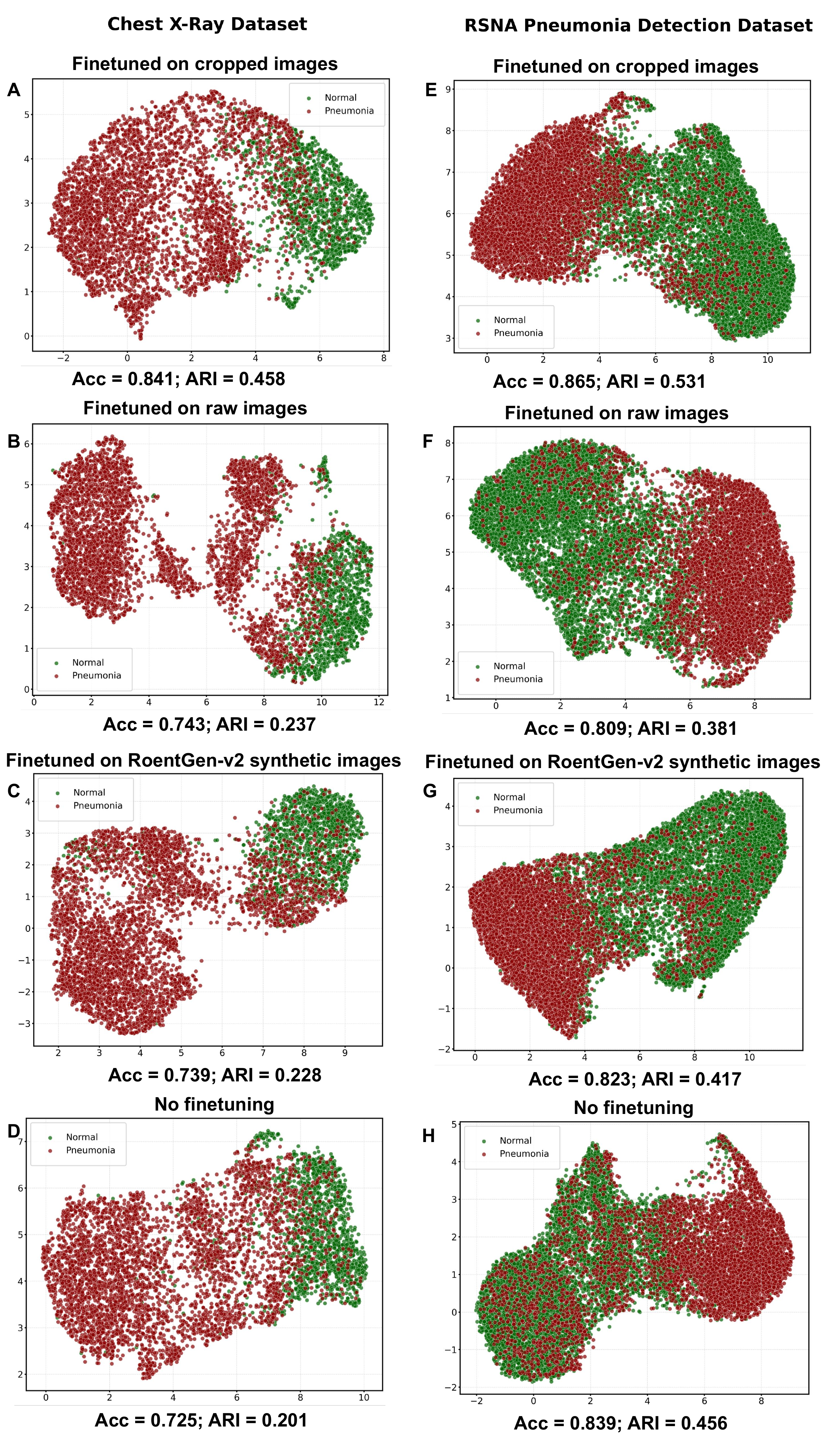}}
\caption{Uniform Manifold Approximation and Projection (UMAP) visualization of features extracted from the classifiers fine-tuned on different datasets. For each feature set, we performed K-means clustering and reported the resulting classification Accuracy (Acc) and adjusted Rand index (ARI). We conducted evaluation separately on the Chest X-Ray dataset (n=5856) (A-D) and the 2018 RSNA Pneumonia Detection Challenge dataset (n=14 863) (E-H).}
\label{umap}
\end{figure}

\begin{figure}[H]
\centerline{\includegraphics[width=\linewidth]{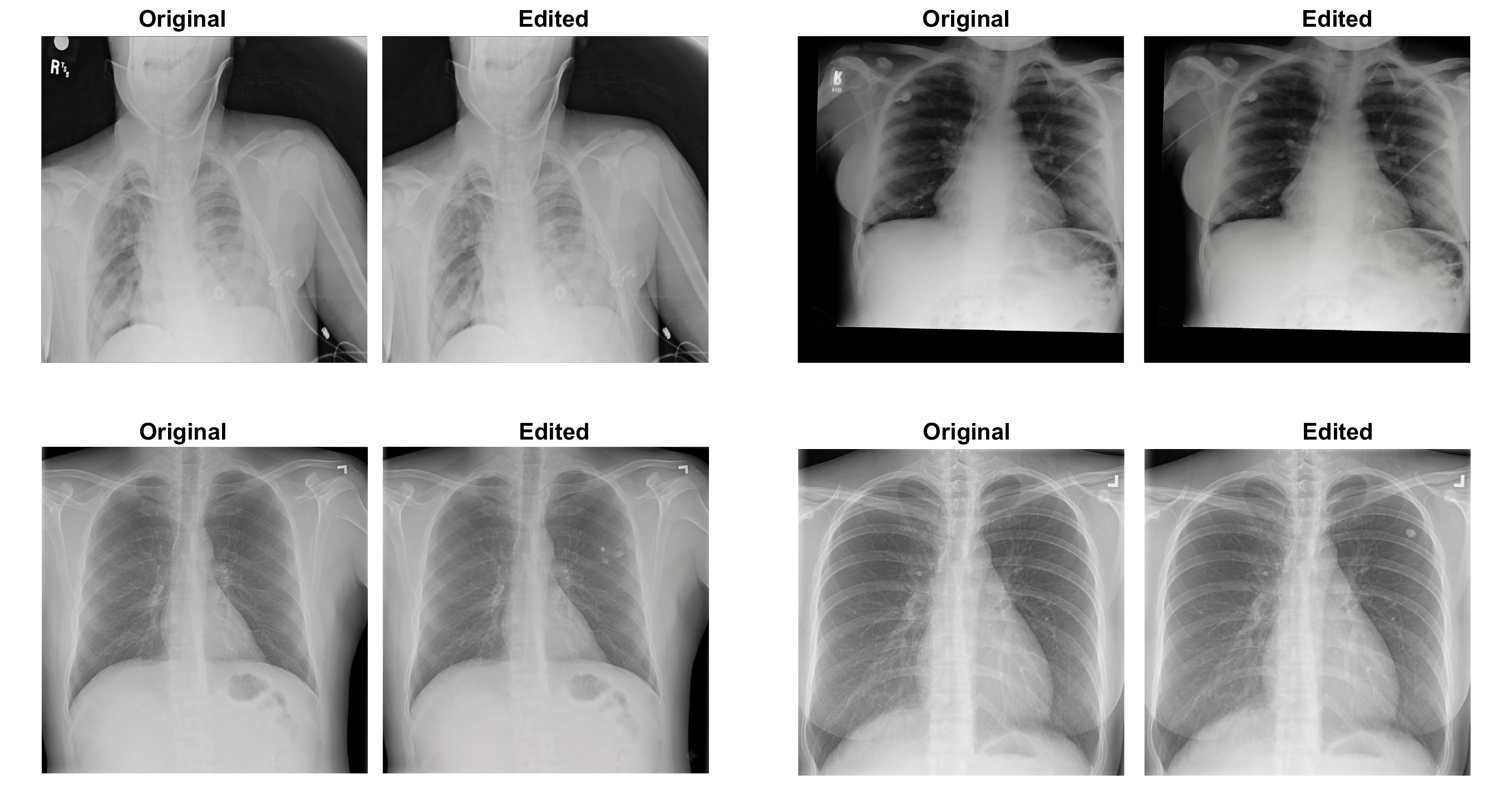}}
\caption{Examples of RSNA chest X-ray image editing. The top row shows two examples of removing image markers and labels from the original images, while the bottom row shows two examples of adding a mild calcified granuloma. In each pair, the left image is the original image and the right image is the edited version.}
\label{cxr_edit}
\end{figure}

\begin{figure}[H]
\centerline{\includegraphics[width=\linewidth]{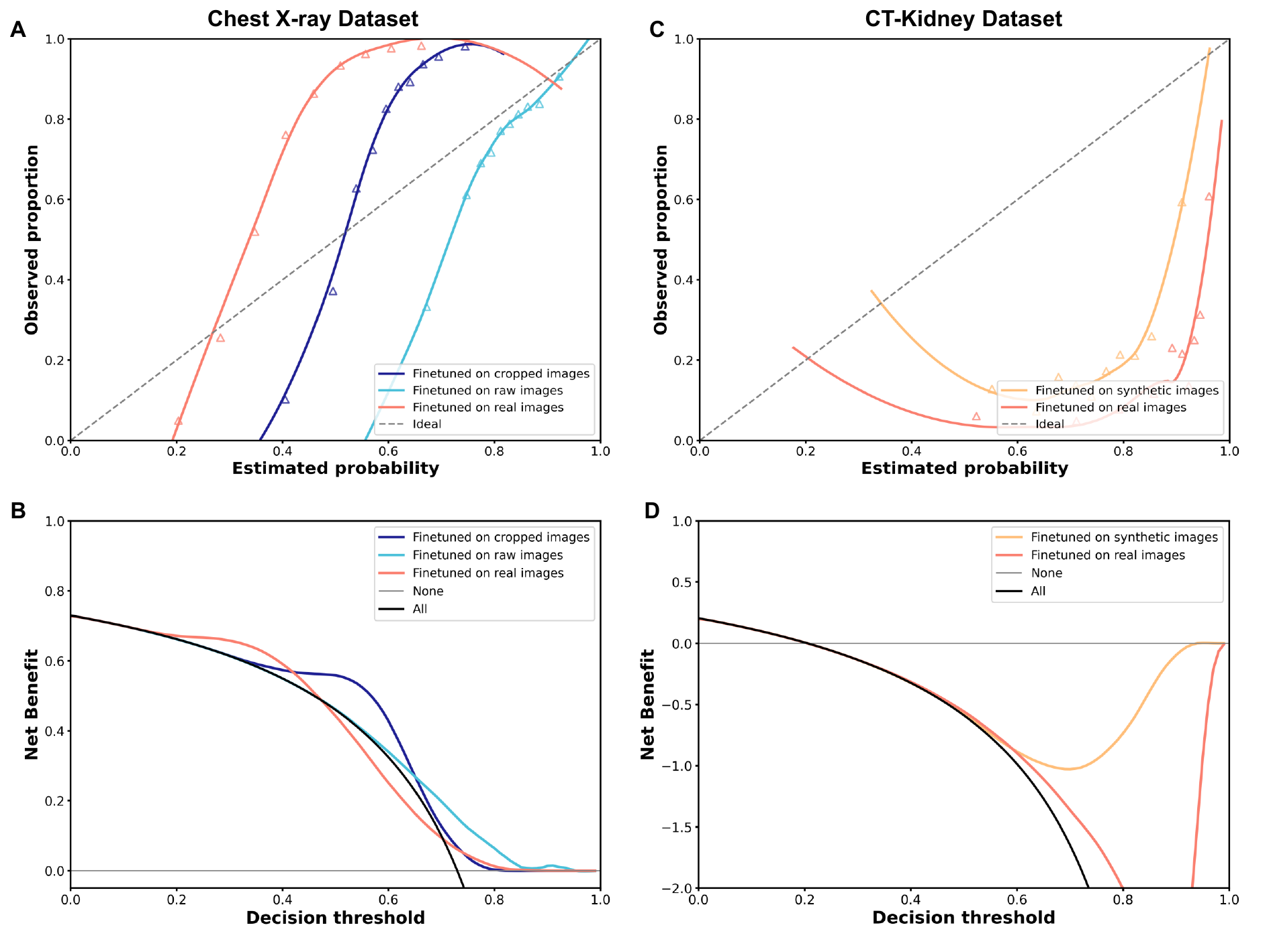}}
\caption{Calibration and decision-curve analyses for CXR and CT diagnostic tasks. In the calibration plots, solid curves represent locally smoothed calibration estimates, open triangles represent grouped estimates from ten approximately equal-sized probability groups, and the grey dashed diagonal denotes perfect calibration.
Smoothed calibration plots (A and C) compare model estimated probabilities with observed event proportions. In the decision curves, colored curves represent smoothed model net benefit, while the grey horizontal line and black curve denote the treat-none and treat-all strategies, respectively. A model provides greater decision-analytic utility where its curve lies above both reference strategies. Decision curves (B and D) show the corresponding net benefit across decision thresholds. Panels A-B show pneumonia recognition on the Chest X-Ray dataset, comparing classifiers trained on cropped Nano Banana-generated CXRs, raw Nano Banana-generated CXRs, and size matched real CXRs from CheXpert. Panels C-D show renal calculus detection on the CT-Kidney dataset, comparing classifiers trained on generated CT images and real CT images. For each training condition, predicted probabilities were averaged across repeated seed runs before plotting.}
    \label{DCA}
\end{figure}

\begin{figure}[H]
\centerline{\includegraphics[width=\linewidth]{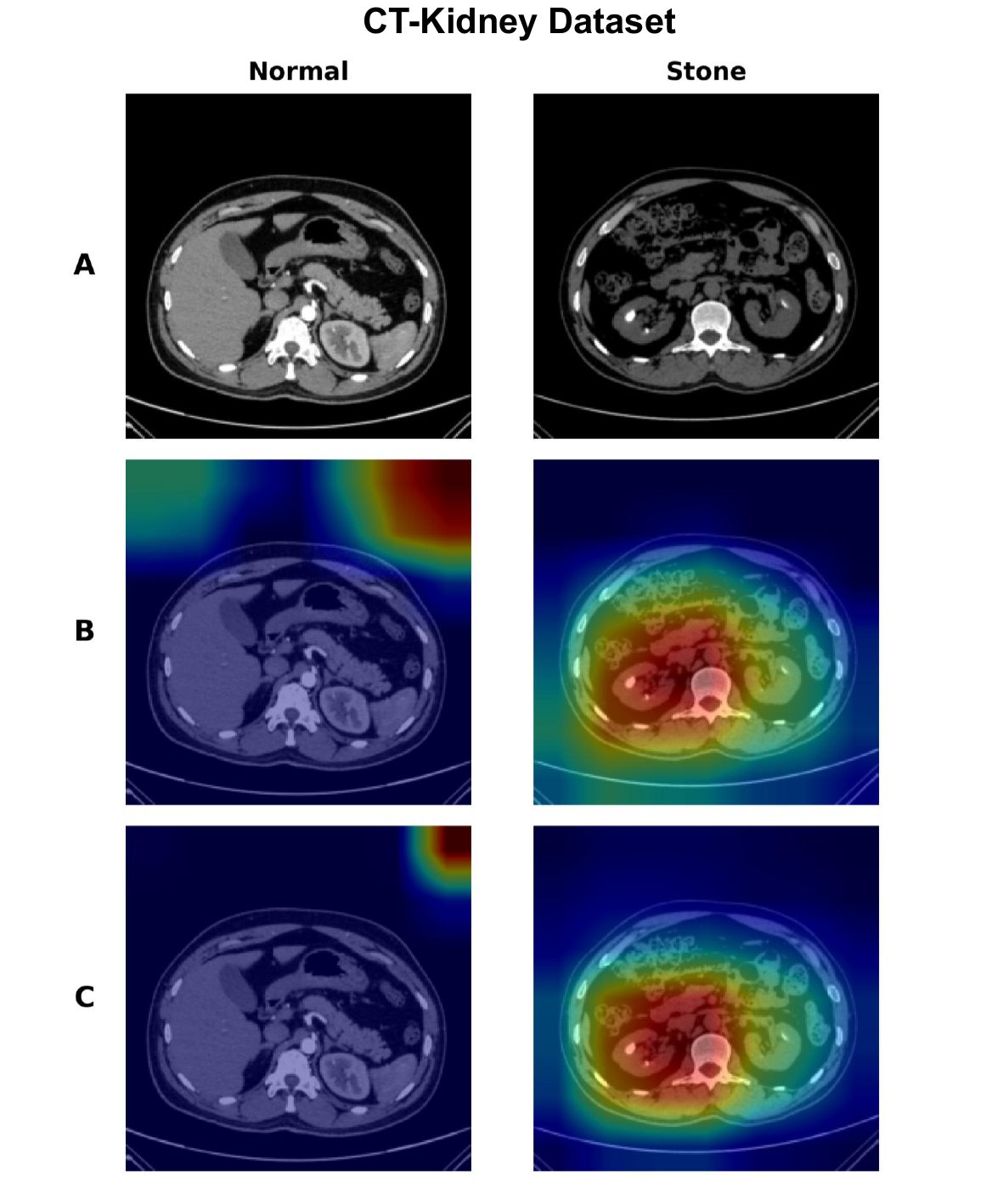}}
\caption{Grad-CAM visualization of the feature map. A: original image; B: fine-tuned on processed CT scans generated by Nano Banana; and C: on real CT scans.}
\label{gradcam_ct}
\end{figure}

\begin{figure}[H]
\centerline{\includegraphics[width=\linewidth]{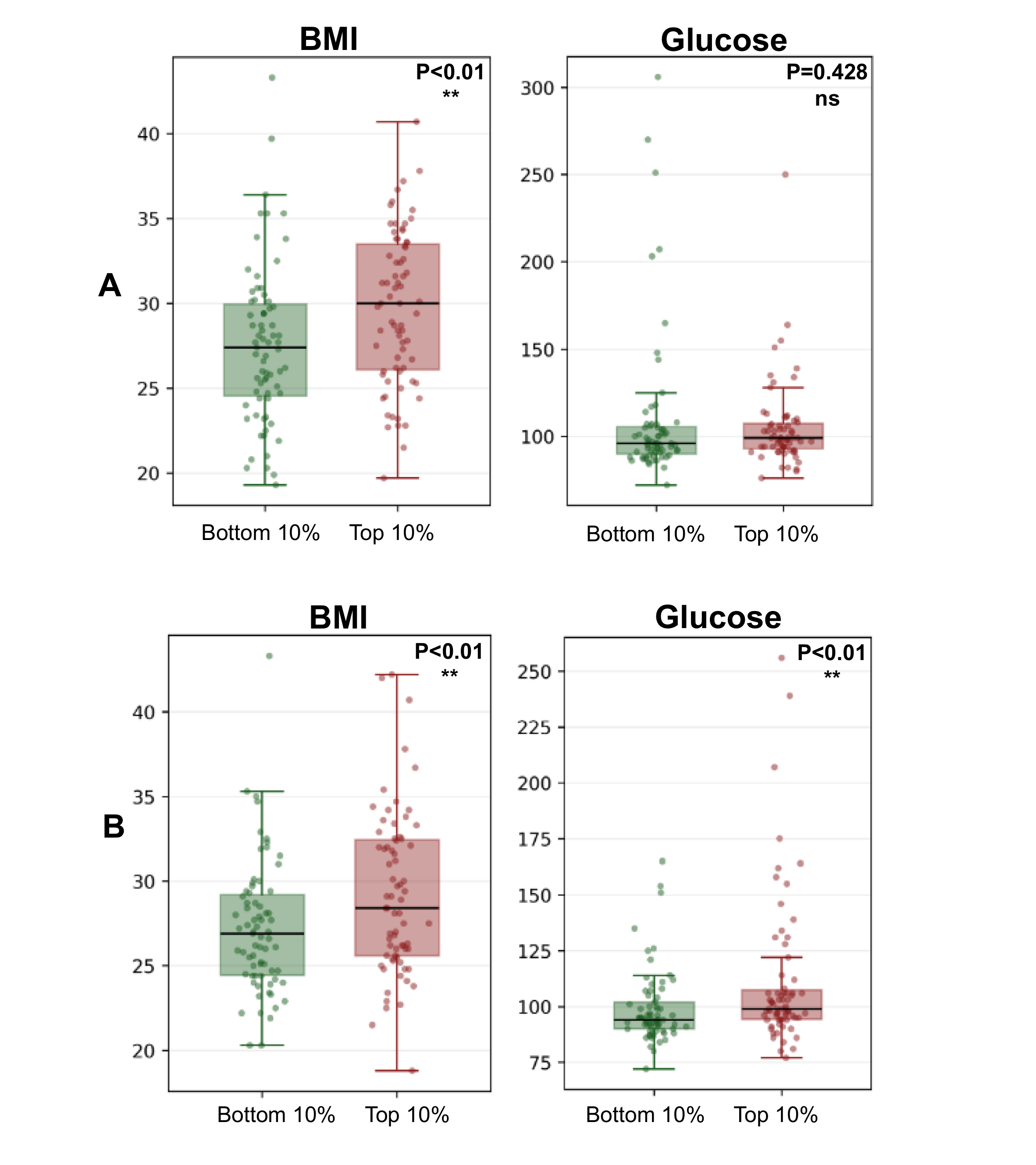}}
\caption{Clinical characteristics of participants with low and high carotid ultrasound-derived ordinal age index. (A) Model trained using HPP data alone. (B) Model trained using HPP data supplemented with synthetic images generated for the 40–44 age group. Results are presented as box plots with individual participant values, with the center line indicating the median and boxes indicating the interquartile range. Welch t-test was used for statistical significance test, and the corresponding P values are shown.}
\label{measure_comp}
\end{figure}

\begin{figure}[H]
\centerline{\includegraphics[width=\linewidth]{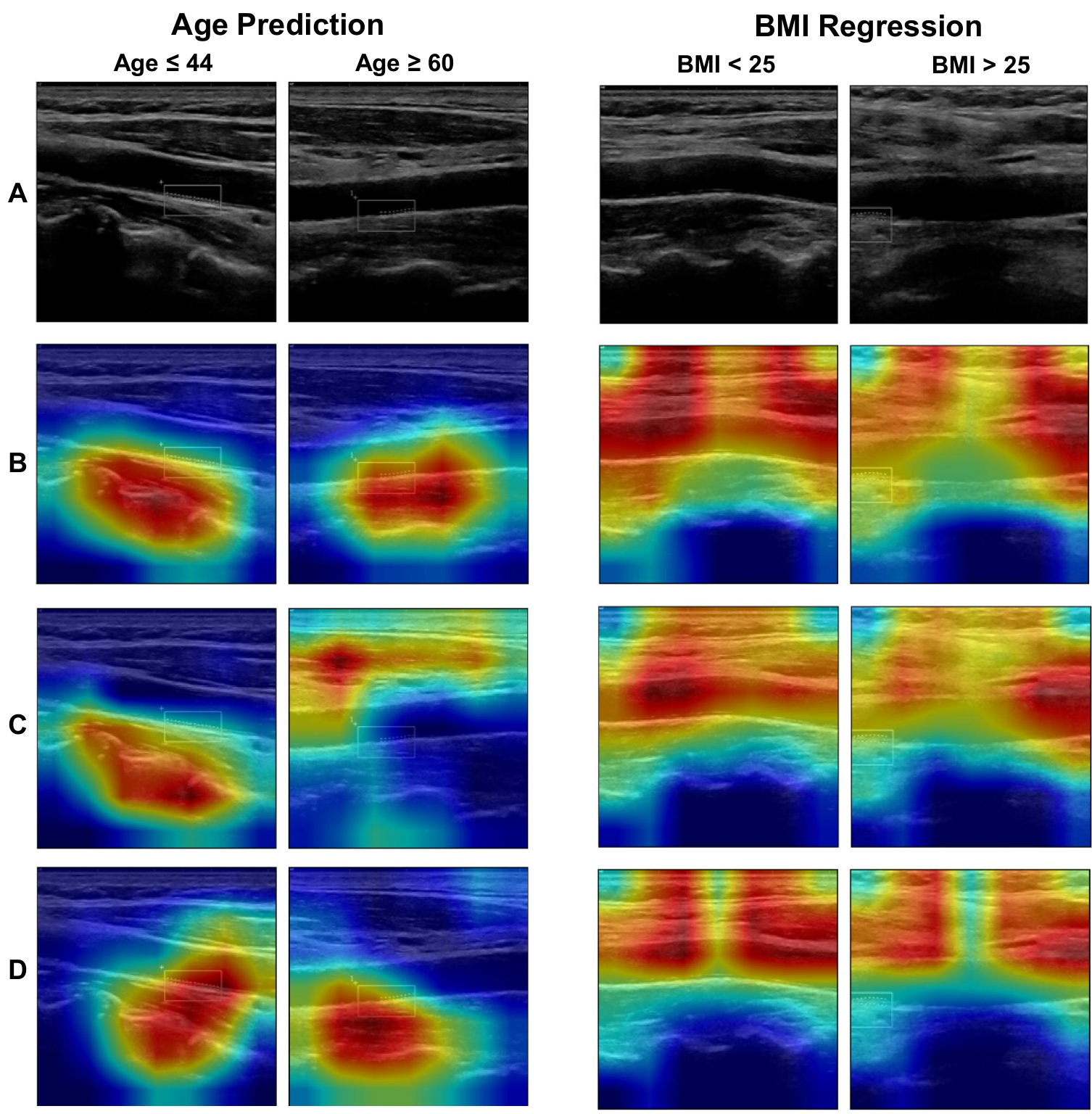}}
\caption{Grad-CAM visualizations of the feature maps for models trained with: (B) mixed data (HPP data and cropped synthetic images); (C) mixed data (HPP data and raw synthetic images); and (D) HPP data alone.}
\label{gradcam_US}
\end{figure}

\begin{figure}[H]
\centerline{\includegraphics[width=\linewidth]{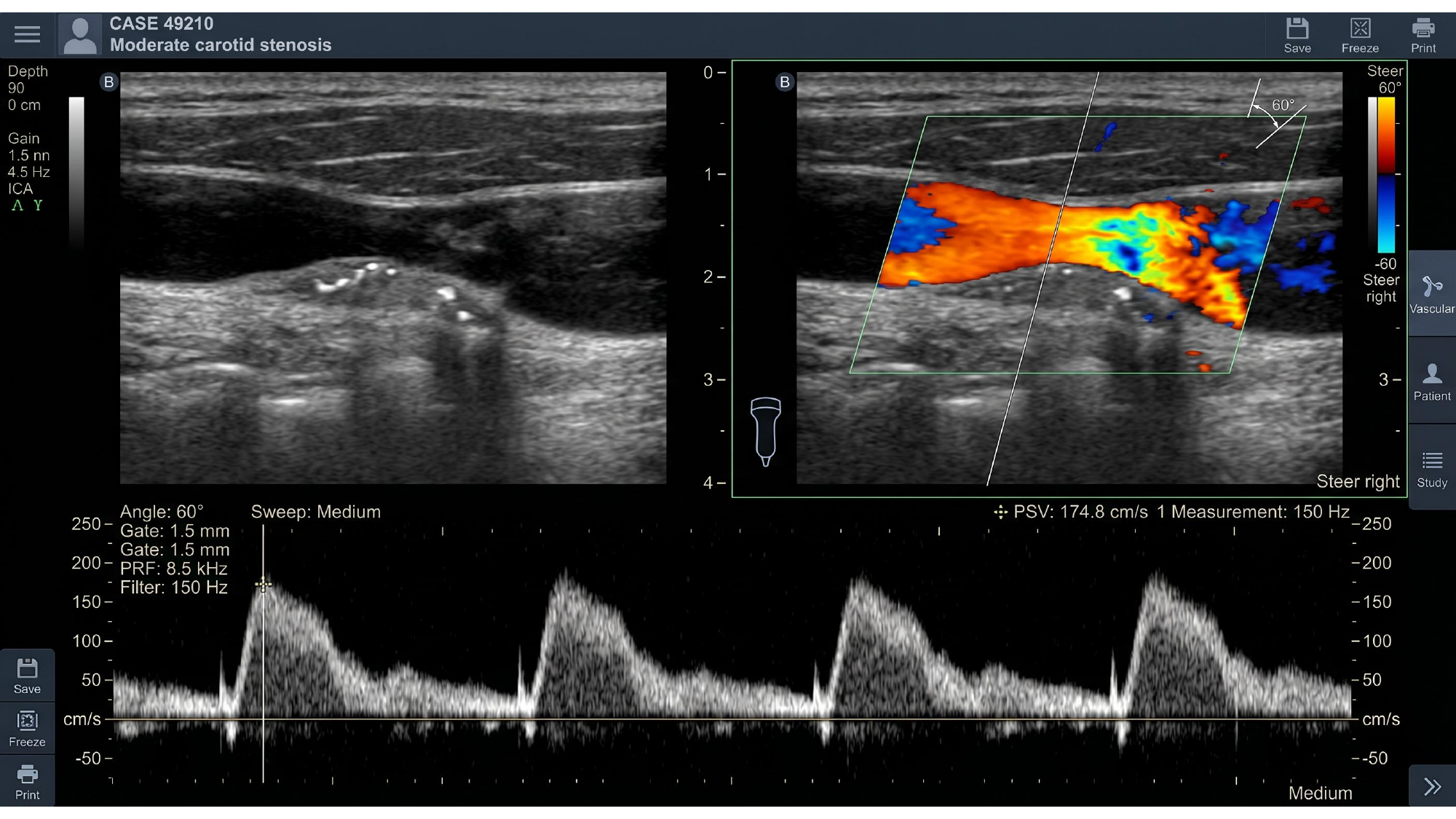}}
\caption{Example of a generated carotid Doppler ultrasound image showing B-mode anatomy, color-flow mapping, and the corresponding spectral Doppler waveform in moderate carotid stenosis.}
\label{doppler_US}
\end{figure}

\begin{figure}[H]
\centerline{\includegraphics[width=\linewidth]{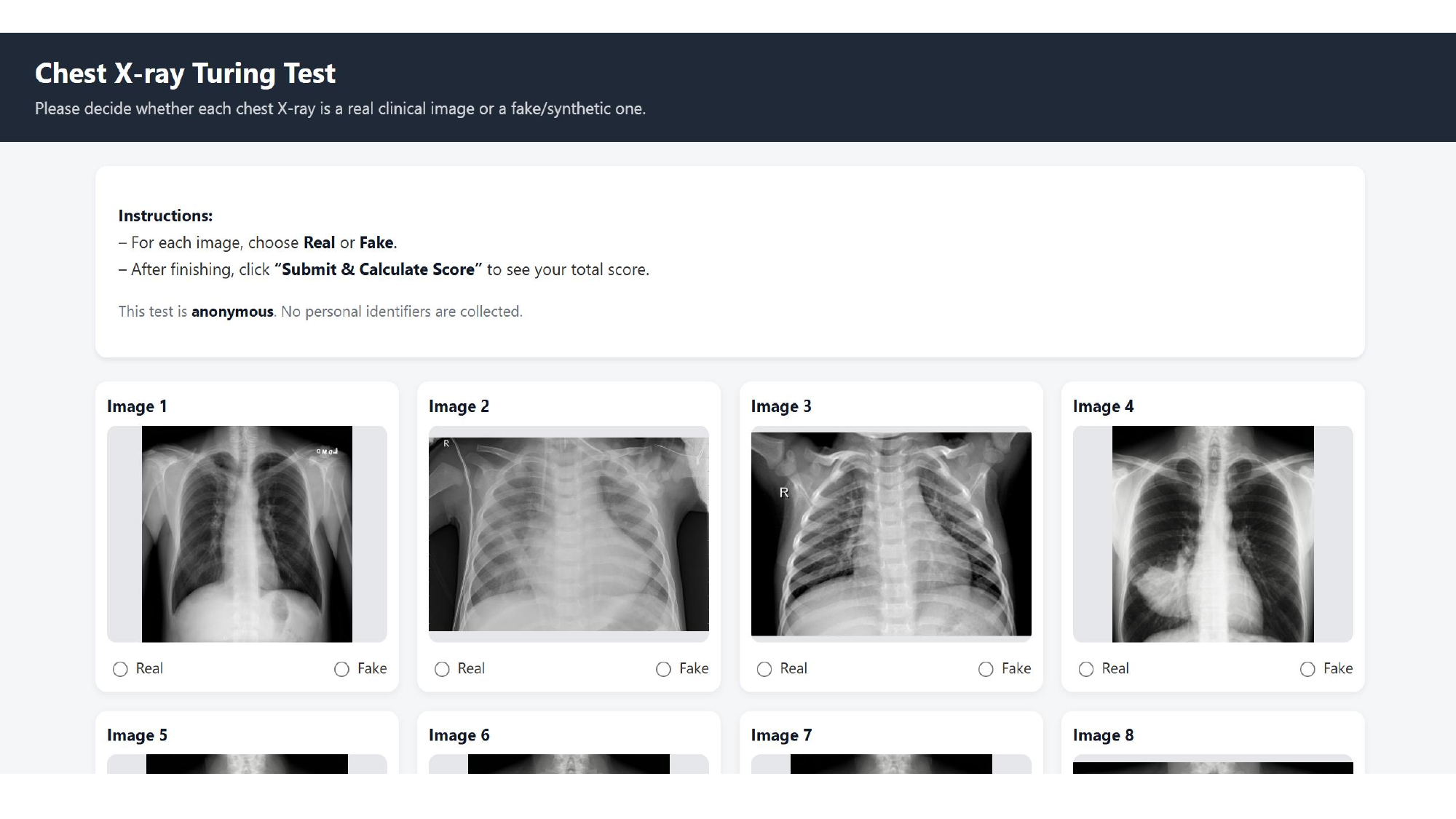}}
\caption{Screenshot of the developed website for CXR Turing Test. 50 synthetic CXRs and 50 real CXRs randomly selected from Chest X-ray Dataset and RSNA Pneumonia Dataset were mixed and shuffled for test. The participant was not informed of the exact number of synthetic and real images included in the test.}
\label{turing}
\end{figure}

\begin{figure}[H]
\centerline{\includegraphics[width=\linewidth]{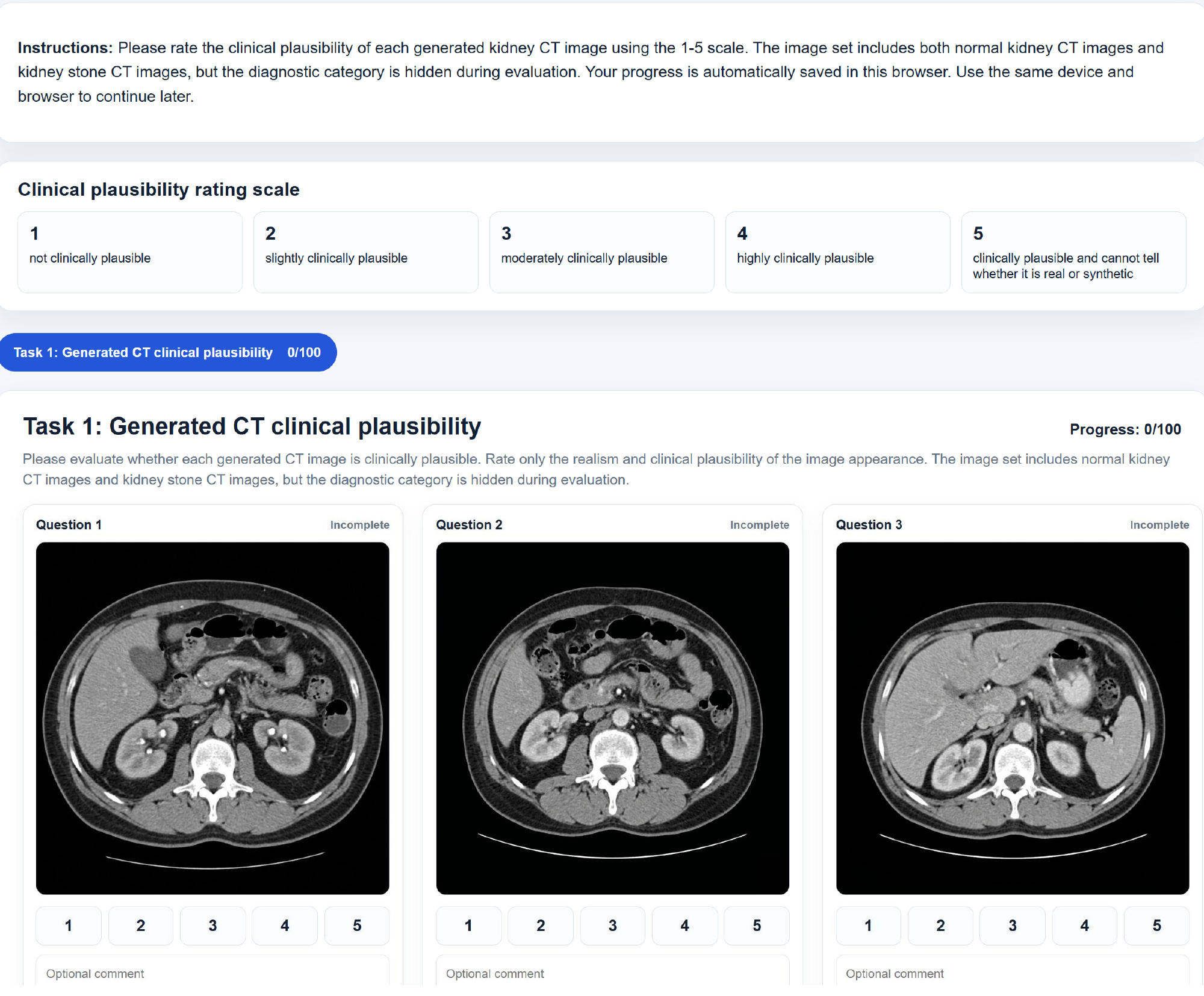}}
\caption{Screenshot of the developed website for the CT clinical plausibility test. The expert rated the clinical plausibility of generated kidney CT images on a 1–5 scale.}
\label{Q_CT}
\end{figure}

\begin{figure}[H]
\centerline{\includegraphics[width=\linewidth]{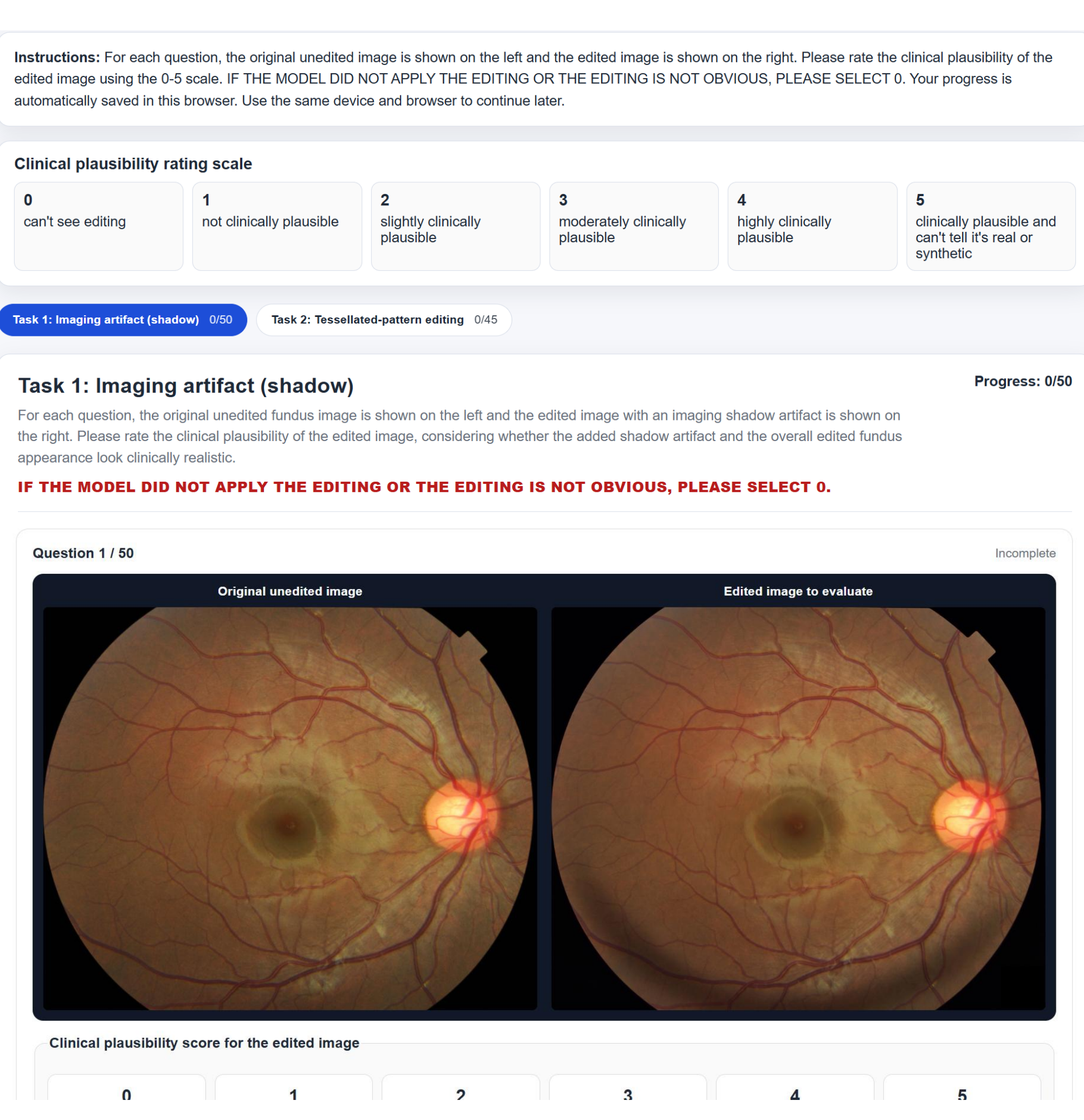}}
\caption{Screenshot of the developed website for the fundus image editing clinical plausibility test. The expert rated edited fundus images on a 0–5 scale across two tasks involving different types of image editing: imaging artifact editing and tessellated-pattern editing.}
\label{Q_fundus}
\end{figure}

\begin{figure}[H]
\centerline{\includegraphics[width=\linewidth]{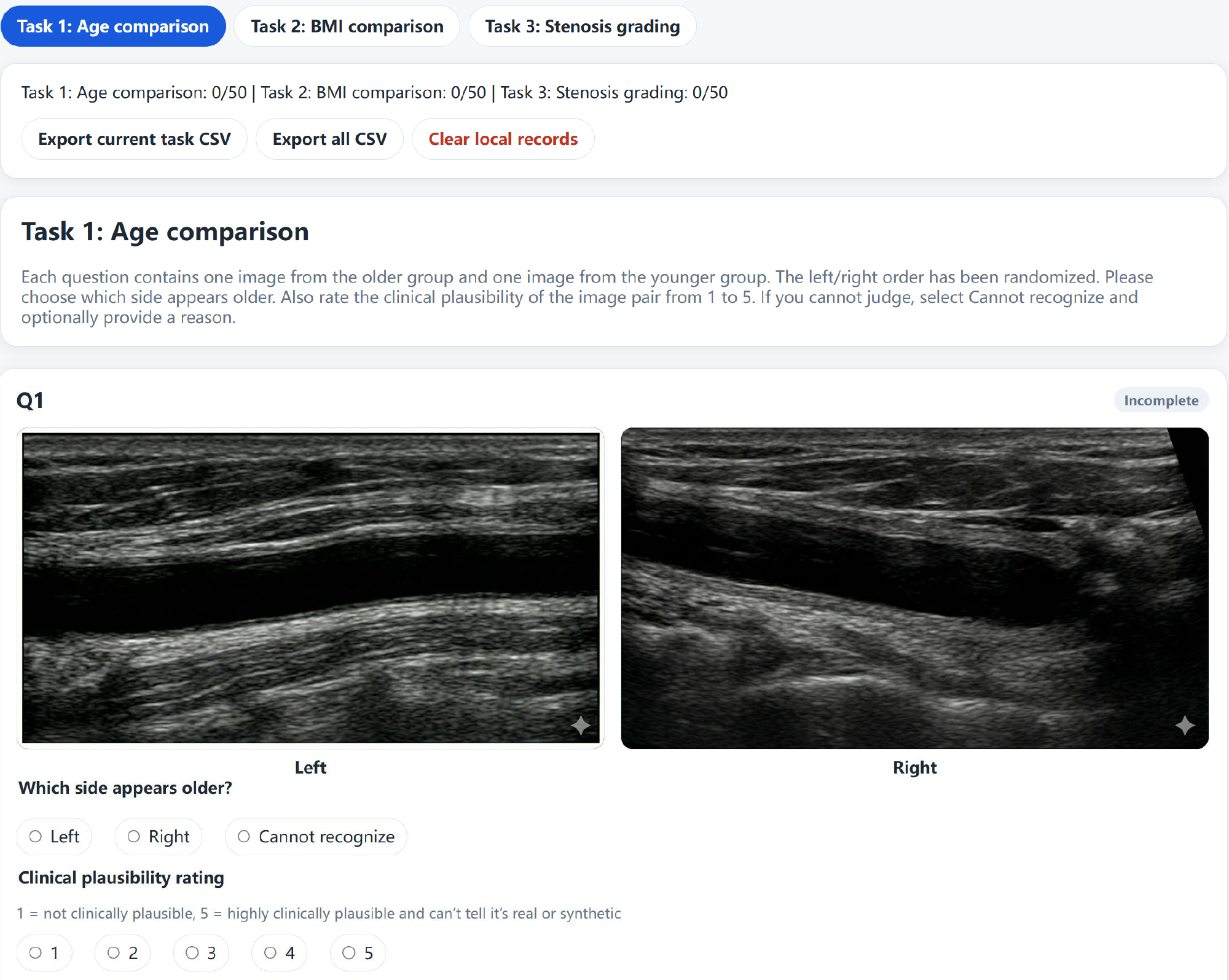}}
\caption{Screenshot of the developed website for the carotid ultrasound reader study. The study included three tasks: age comparison, BMI comparison, and carotid stenosis grading with Doppler images. The expert also rated the clinical plausibility of the ultrasound images on a 1–5 scale.}
\label{Q_carotid}
\end{figure}

\begin{figure}[H]
\centering
\includegraphics[height=0.85\textheight, keepaspectratio]{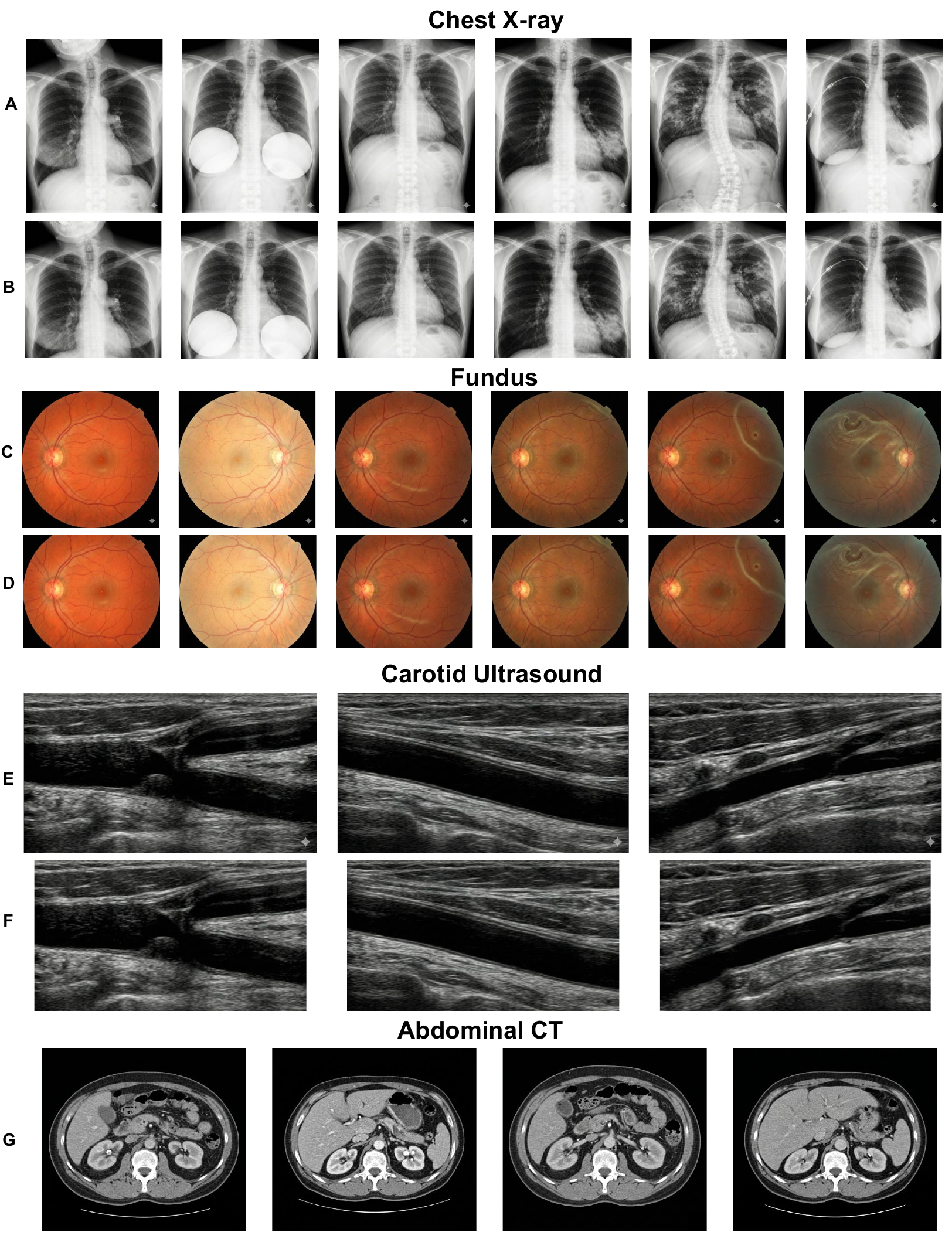}
\caption{Representative raw and cropped images generated by Nano Banana across four imaging modalities. For each modality, the top row (A, C, E) shows raw images and the bottom row (B, D, F) shows the corresponding cropped images. Rows A--B (chest X-ray): healthy subjects (columns 1--3) and subjects with pneumonia (columns 4--6), illustrating diverse acquisition characteristics, including acquisition-related pose tilt (column 1), breast prostheses (column 2), and suboptimal inspiratory effort (column 3). 
Rows C--D (fundus photograph): normal eyes (columns 1--2), mild RD (columns 3--4), and severe RD (columns 5--6). 
Rows E--F (carotid ultrasound): subjects aged $\geq 60$ years (column 1), 
subjects aged $<45$ years (column 2), and individuals with BMI $17 \leq \mathrm{BMI} \leq 25$ (column 3).
Row G (abdominal CT): examples with renal calculi (columns 1--2) and without renal calculi (columns 3--4).}
\label{examples}
\end{figure}

\backmatter

\bibliography{sn-bibliography}% common bib file

\end{document}